%% file: acl_latex.tex
\documentclass[11pt]{article}

\usepackage{acl}

\usepackage{times}
\usepackage{latexsym}
\usepackage{booktabs}
\usepackage[most]{tcolorbox}
\usepackage{enumitem}
\usepackage{pifont}
\usepackage{booktabs}
\usepackage{adjustbox}
\usepackage{tabulary}
\usepackage{enumitem}

 \usepackage{amsmath} 
 \usepackage{makecell}
\usepackage[table]{xcolor}
\usepackage{longtable}

\usepackage{graphicx}
\usepackage{rotating}
\usepackage{array}

\newcommand{\cmark}{\ding{51}}
\newcommand{\xmark}{\ding{55}}
\newcommand{\pmark}{\raisebox{0.2ex}{\scriptsize$\triangle$}}
\newcommand{\nr}{\textsc{NR}}
\usepackage[T1]{fontenc}

\usepackage[utf8]{inputenc}

\usepackage{microtype}

\usepackage{inconsolata}

\usepackage{graphicx}
\usepackage{xspace}

\newcommand{\ours}{\textsc{AgentInteractError}\xspace}

\title{Understanding Communication Failures in Multi-Agent LLM Systems}

\author{
\textbf{Priya Pitre} \quad
\textbf{Jun Hou} \quad
\textbf{Yi Fang} \\
\textbf{Naren Ramakrishnan} \quad
\textbf{Xuan Wang} \\
Virginia Tech \\
Blacksburg, VA 24061, USA
}

\begin{document}
\maketitle
\begin{abstract}
Multi-agent LLM systems aim to improve reasoning through debate, evidence sharing, and mutual correction, but interaction can also introduce failures that final accuracy alone cannot explain. Agents may converge too early, ignore dissent, suppress correct minority answers, or preserve a wrong framing across rounds. We present a corpus-grounded taxonomy of communication failures, built from 3,211 failed multi-round trajectories across medical QA, coding, finance, multi-hop reasoning, ethics, and embodied text tasks. Using this taxonomy, we analyze failure distributions across datasets and system designs, round-level error dynamics, recovery hardness, majority transitions, minority-signal suppression, and self-reported confidence. We find that failures are dominated by recovery/correction and consensus/decision problems rather than isolated first-round mistakes, with many emerging after agents first observe peer responses. Based on these findings, we recommend stronger disagreement handling, evidence verification, minority-signal preservation, and recovery checks before finalization.
\end{abstract}

\input{Sections/01Introduction}

\input{Sections/02RelatedWorks}

\input{Sections/03Methodology}
\input{Sections/04Taxonomy}

\input{Sections/05Results}

\input{Sections/06Suggestions}
\input{Sections/limitations.tex}

\bibliography{custom}

\appendix
\label{sec:appendix}

\input{Sections/Appendix}

\end{document}

%% file: Sections/01Introduction.tex
\section{Introduction}

\begin{figure*}
  \centering
  \includegraphics[width=\textwidth]{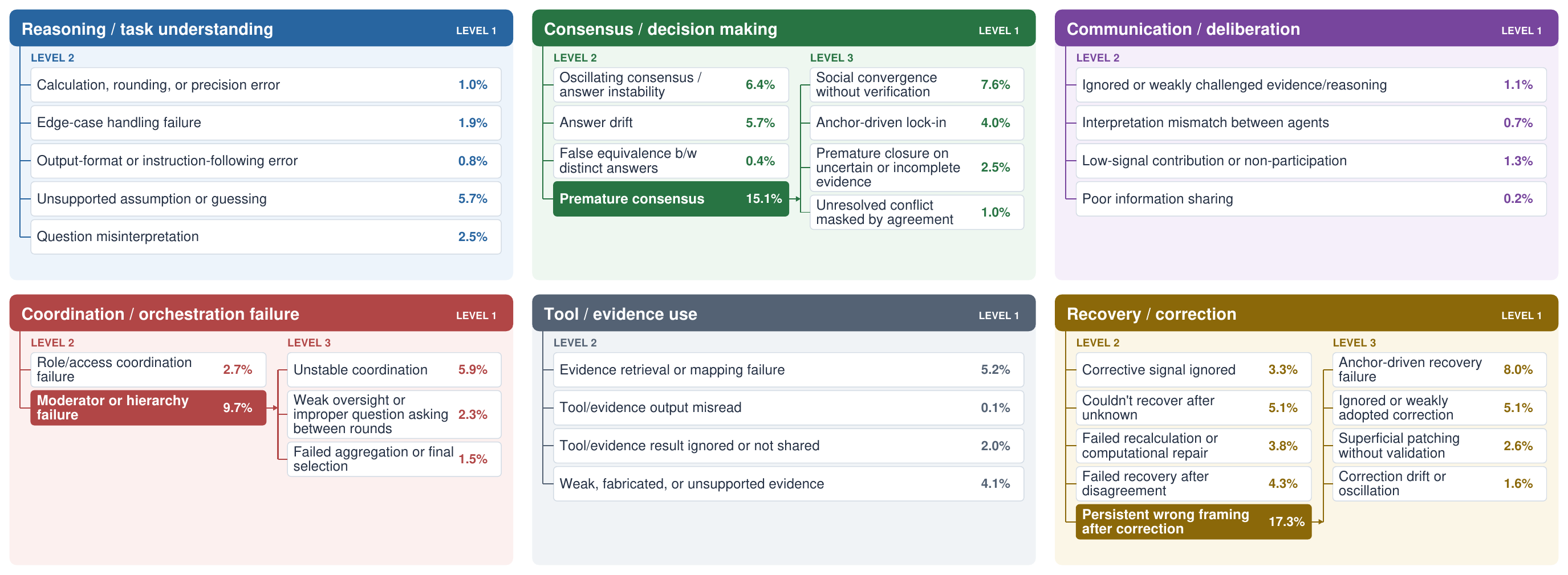}
  \caption{Final taxonomy of multi-agent communication failures, organized into parent categories and fine-grained error types.}
  \label{fig:taxonomy}
\end{figure*}

Multi-agent LLM systems are motivated by a simple but powerful premise: failures of a single model may be easier to catch when several agents divide work, exchange evidence, and challenge one another before producing a final answer. This premise now appears across both research and deployment. Multi-agent debate uses interacting models to critique and revise rationales \cite{du2024improving}; discussion-based approaches use diverse models to reconcile disagreements \cite{chen2024reconcile}; sampling-and-voting methods instantiate multiple agents to improve robustness; and production systems such as Amazon Bedrock support supervisor agents that coordinate specialized collaborators \cite{amazonbedrock}. Across these settings, communication is the central mechanism through which agents decompose tasks, transfer evidence, resolve disagreement, incorporate tool outputs, and decide when to stop.

However, the same communication process that makes multi-agent systems appealing also creates a new failure surface. Discussion can repair individual mistakes, but it can also amplify them: correct answers may be abandoned after peer influence \citep{wynn2025talk}, failures may arise from system design, inter-agent misalignment, and weak verification \citep{cemri2026multi}, and the responsible agent or decisive failure step can remain difficult to localize even after the system fails \citep{zhang2025agent}. These findings suggest that multi-agent errors are not simply individual model errors repeated in parallel, but interaction-level failures produced through communication, authority, aggregation, and recovery.

Existing taxonomies and diagnostic benchmarks provide important starting points, but they leave the communication process in multi-agent discussion under-specified. \citet{cemri2026multi} identifies broad sources of failure such as system design, inter-agent misalignment, and weak verification; \cite{deshpande2025trail} and \cite{barke2026agentrx} analyze agentic traces and critical failure steps; and attribution-focused work such as \cite{chen2026seeing}, \cite{yu2025correct} and \cite{ma2026demystifying} studies which agent, step, or platform component contributed to failure. 

These works show the value of trace-level diagnosis, but they do not make interaction itself the object of analysis. This matters because multi-agent failures are often produced between agents rather than inside one agent or component: a correct minority answer can be abandoned, a weak claim can become dominant through repetition, and a useful correction can disappear if it is not taken up before finalization. Final accuracy only tells us that the system failed; it does not show whether the failure came from initial reasoning, peer influence, failed recovery, or aggregation. A communication-centered taxonomy makes these interaction patterns measurable and comparable across datasets, models, agent counts, and orchestration settings.

To address this, we construct a corpus-grounded taxonomy of multi-agent communication failures and use it to study how errors emerge, spread, and recover during deliberation. Our corpus spans ten domains, including medical QA, coding, finance, multi-hop reasoning, ethics, and embodied text environments, while varying model family, agent count, tool access, composition, and orchestration style. We focus on incorrect multi-round trajectories, using correct multi-round trajectories as a contrast set, to analyze when errors first appear, who introduces them, whether recovery is attempted, and how group answers change across rounds. To our knowledge, this is the first communication centric taxonomy focusing on multi-agent LLM debates. 

Rather than deriving categories from an unconstrained LLM prompt, we build the taxonomy from trajectory evidence, inspired by \citet{kargupta-etal-2025-taxoadapt}. Each trace is summarized into a compact evidence card, then mapped into diagnostic categories with an explicit option for unsupported cases. The taxonomy is iteratively expanded when dense or unmapped evidence reveals coherent new failure types.

Our analysis shows that multi-agent failures are dominated by recovery and consensus problems rather than isolated first-round mistakes: recovery/correction failures appear in 94.9\% of failed trajectories and consensus/decision failures in 83.4\%. Round 2 is a critical point where many failures emerge after agents first observe peer responses, and correct minority signals are often suppressed rather than amplified, especially in larger groups. Finally, we show that the taxonomy can support real-time early-warning monitors for persistent wrong framing, premature consensus, and oscillating consensus, achieving very high recall before the final answer is produced. Based on these findings, we propose recommendations for designing more reliable multi-agent systems.

%% file: Sections/02RelatedWorks.tex
\section{Related Work}

\paragraph{Multi-agent Taxonomy:} Multi-agent LLM systems distribute reasoning across agents that debate, specialize, verify, vote, or use tools for one another. Multi-agent debate, ReConcile, More Agents Is All You Need, and AutoGen share the premise that rationales or aggregation can expose errors a single model might miss \cite{du2024improving, li2024more, wu2024autogen}, but recent work shows these gains are conditional: \citet{cemri2026multi}  finds setup-dependent failures in system design, inter-agent misalignment, and task verification,  while \citet{wynn2025talk} shows that debate can reduce accuracy when agents abandon correct answers after peer responses. Related taxonomies and attribution benchmarks, including \citet{cemri2026multi, zhu2025llm, barke2026agentrx, yu2025correct, kong2025aegis} platform root-cause and lifecycle studies while \citet{bi2026diagnosing, zhang2025agent, chen2026seeing, in2026rethinking} localize, transfer, or generate failure patterns across traces. Our work focuses on automatically constructed communication centered taxonomy. Additional details are in Appendix B.

\paragraph{Multi-agent error detection:} Existing works use failure signals for detection and mitigation. AgentForesight performs online auditing during an unfolding trajectory~\cite{zhang2026agentforesight}; Problem Drift introduces DRIFTJudge and DRIFTPolicy for debate drift detection and feedback~\cite{becker2026stay}; From Spark to Fire models error cascades and mitigates them with message-level governance~\cite{xie2026spark}; InferAct detects misaligned actions before execution~\cite{fang2025preemptive}; and AgentDebug and AgentFixer connect detection to feedback, repair recommendations, and validation~\cite{zhu2025llm,mulian2026agentfixer}. These systems move beyond post-hoc explanation, but their signals are often task-, action-, or drift-specific, and many intervene only after failure is visible. Our trigger system is real-time and taxonomy-guided: after each round, it uses communication-error labels to estimate whether continued deliberation will recover or amplify an error. Because the taxonomy spans domains, tools, setups, and multiple errors per trace, it supports propagation-aware intervention rules beyond final correctness or generic judge scores.

%% file: Sections/03Methodology.tex
\section{Methodology}

Our methodology has three stages: collecting multi-agent trajectories, constructing a corpus-grounded communication-error taxonomy, and applying the taxonomy to a downstream trigger-error detection task.

\paragraph{Multi-Agent Trajectory Collection}

We collected multi-agent debate traces from eight datasets spanning financial QA, multi-hop QA, medical QA, ethical judgment, multiple-choice reasoning, and code generation: FinQA \cite{chen2021finqa}, HotpotQA \cite{yang2018hotpotqa}, MBPP \cite{austin2021program}, MMLU-Pro \cite{wang2024mmlu}, MedQA \cite{yao2024medqa}, MuSiQue \cite{trivedi2022musique}, Ethics \cite{hendrycks2020aligning}, and LiveCodeBench \cite{jain2025livecodebench}. Agents were drawn from open-weight models including LLaMA-3.2-3B, Mistral-7B, Qwen-14B, LLaMA-70B, GPT-OSS-20B, GPT-OSS-120B, and Gemma-12B, with assignments varying by experiment variant. Across all experiment files, we ran 7,300 task instances, 3,543 of which ended incorrectly. We focus on incorrect trajectories with more than one debate round, yielding 3,211 traces for taxonomy construction. For each experiment variant, we used a fixed 100-example sample where available, sampled with seed 42. Accuracy and tested settings are reported in Table \ref{tab:accuracy_by_agent_configuration}; full protocol and dataset-specific details are in Appendix \ref{sec:experimental_setting}.

\paragraph{Taxonomy Construction.}
We followed a TaxoAdapt-inspired \cite{kargupta-etal-2025-taxoadapt} procedure combining top-down initialization with bottom-up expansion. Each failed multi-round trajectory was converted into a compact card containing agent trajectories, round-level errors, failure summaries, candidate tags, and supporting evidence. GPT-5.4-nano handled the main generation and mapping steps, including two-pass card generation: summarizing failure behavior, then producing 3--8 reusable tags with evidence and justification.

\begin{itemize}[leftmargin=*]
\item \textbf{Initialize broad dimensions.} We specified six top-level dimensions: reasoning/task-understanding, communication/deliberation, consensus/decision, recovery/correction, coordination/orchestration, and tool/evidence-use failure. Lower-level labels were induced from the corpus.

\item \textbf{Map and deduplicate tags.} Candidate tags were mapped to broad dimensions, leaving weak fits unmapped and trajectories multi-label. Near-duplicates were merged using text normalization, stopword removal, token Jaccard similarity, containment, and RapidFuzz token-set ratio. We retained the shortest clear representative tag (earliest row breaking ties), along with identifiers, duplicate tags, evidence, justifications, counts, and row indices.

\item \textbf{Seed child labels.} For each dimension, GPT-5.4-nano proposed 5--6 child labels from 50 sampled representative tags. We manually removed redundancy and improved interpretability. GPT-5.4-nano also handled broad-category assignment, parent-label cleanup, and broad-label checking/correction.

\item \textbf{Expand width from unmapped tags.} Unmapped tags after top-down mapping were treated as coverage gaps and clustered using only representative-tag text. We used BAAI/bge-base-en-v1.5 embeddings, optional PCA (at most 50 dimensions), and agglomerative clustering with cosine distance and average linkage. Candidate cluster counts (5, 10, 15, 20, 25, 50, 100) were evaluated by cosine silhouette score and centroid-nearest representatives. GPT-4.1-mini reviewed clusters for existing or new labels, or left them unmapped; proposed labels were checked and consolidated.

\item \textbf{Expand depth for large leaves.} Leaves exceeding 600 representative rows were subdivided using the same procedure, testing cluster counts of 3, 4, 5, 6, 8, 10, 12, 15, and 20. These included premature consensus, moderator/hierarchy failure, and persistent wrong framing after correction, with GPT-4.1-mini performing depth classification.

\item \textbf{Iterate and preserve evidence.} We repeated mapping, clustering, redundancy checks, consolidation, and depth expansion until reaching depth 3 where needed, with no leaf above the threshold. Row-level outputs retain \texttt{broad\_category}, \texttt{child}, \texttt{mapped\_label}, \texttt{representative\_tag}, \texttt{unique\_identifier}, \texttt{evidence}, and \texttt{justification}; trajectory-level outputs aggregate labels while preserving evidence links.
\end{itemize}

\begin{figure}
  \centering
  \includegraphics[width=\linewidth]{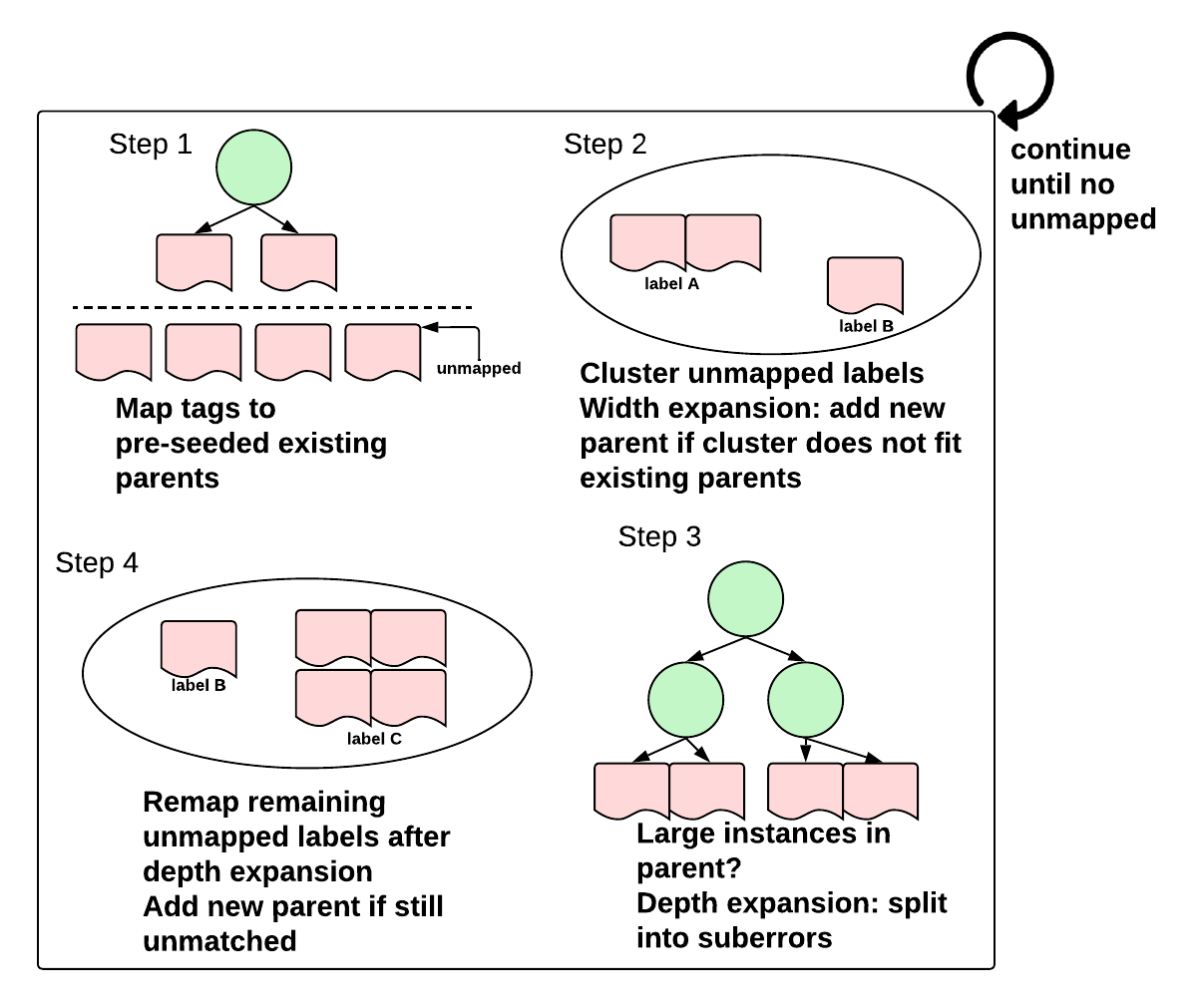}
  \caption{Our automatic taxonomy creation methodology}
  \label{fig:taxonomy}
\end{figure}

\paragraph{Human Annotation and Taxonomy Validation.} We sampled 200 random failed trajectories, each annotated by three annotators. First, annotators performed open-ended annotation, describing what went wrong without seeing our taxonomy. We then mapped these descriptions to taxonomy labels to measure coverage. Second, annotators were shown the taxonomy and assigned all applicable labels. We report inter-annotator agreement and compare automatic labels against adjudicated human annotations.

\paragraph{Contrast Set.}
To identify which errors distinguish successful from failed debates, we sample 500 correct multi-round trajectories from the same six non-coding datasets as the failed trajectories. We use an outcome-blind annotation prompt that identifies errors from trace evidence without knowing the final outcome, then maps them to our fixed taxonomy. We compare these annotations against failed trajectories, adjusting for differences in dataset distribution.

\paragraph{Trigger Model.}
To demonstrate a downstream use of the taxonomy, we train early-warning models to predict whether an error will appear in the next round. We focus on premature consensus, failed recovery after disagreement, and failure to recover after an unknown answer. We use 304 debates paired by question and correctness, with error onset rounds annotated without access to final outcomes. Our random forest uses answer dynamics, including disagreement, answer changes, oscillation, and confidence, without message text or dataset identity. We evaluate using leave-one-pair-out validation against rule-based baselines and an LLM judge (Appendix~\ref{app:trigger}).

%% file: Sections/04Taxonomy.tex
\section{\ours Taxonomy}

Figure~\ref{fig:taxonomy} shows the final taxonomy. It has six broad parent categories: reasoning/task understanding, communication/deliberation, consensus/decision, recovery/correction, coordination/orchestration, and tool/evidence use. Under these, the taxonomy contains 24 child labels and 11 final leaf labels. 

The labels in the taxonomy are unique but context-dependent. The same behavior can belong to different parts of the taxonomy depending on what it affects in the trajectory. For example, answer drift can be a consensus/decision failure when it makes group agreement unstable, such as when the majority answer changes across rounds or agents converge through momentum rather than verification. However, it can also be a recovery/correction failure when agents had access to a correct answer, correction, or useful evidence, but later discussion pulls them away from it instead of preserving the repair. So the label is not based only on the words in the trace; it depends on the role the behavior plays in the failure.

We also treat each trajectory as multi-error. A failed debate often does not have just one cause: it may start with evidence misuse, then have weak dissent, then end in premature consensus or failed recovery. For this reason, one trajectory can receive multiple labels, each tied to the specific evidence in the trace.

\paragraph{Evaluation of Taxonomy}
Using the metrics introduced by \citet{kargupta-etal-2025-taxoadapt}, we evaluated the final taxonomy along five criteria: path granularity checks whether each child is more specific than its parent, sibling coherence checks whether children under the same parent form a consistent same-granularity set, dimension alignment checks whether nodes remain within the intended failure-mode taxonomy dimension, paper relevance checks whether each node covers at least 5\% of the corpus, and coverage measures whether a parent’s relevant papers are covered by at least one child; the first three semantic metrics were judged with GPT-4.1-mini, while paper relevance and coverage were computed directly from corpus-level taxonomy assignments (Table \ref{tab:taxonomy-eval}).

\begin{table}[!t]
\centering
\scriptsize
\setlength{\tabcolsep}{5pt}
\renewcommand{\arraystretch}{1.05}
\begin{tabular}{@{}llcc@{}}
\toprule
\textbf{Metric} &
\textbf{Evaluation level} &
\textbf{$n$} &
\textbf{Score} \\
\midrule
Path granularity &
Node/path-wise &
37 &
0.946 \\

Sibling coherence &
Level-wise &
9 &
0.778 \\

Dimension alignment &
Node-wise &
43 &
1.000 \\

Relevance &
Node-wise &
66 &
0.712 \\

Coverage &
Level-wise &
32 &
1.000 \\
\bottomrule
\end{tabular}
\caption{\textbf{Final taxonomy evaluation results.}
Scores summarize structural quality, coverage, and relevance checks for the final taxonomy across node-, path-, and level-wise evaluations.}
\label{tab:taxonomy-eval}
\end{table}

%% file: Sections/05Results.tex
\section{Analysis and Discussion}

\paragraph{Human Annotation and Taxonomy Validation.}
We first test whether the taxonomy captures failures that humans identify independently and whether our automatic labels are reliable enough for the analyses that follow. In open annotation, 93.4\% of human-identified errors mapped to a fine-grained taxonomy label and 98.1\% to a parent category. With the taxonomy visible, annotators reached 0.82 Jaccard agreement at the parent level and 0.73 at the fine-grained level; our automatic pipeline reached 0.91 and 0.84 micro-F1 against adjudicated human labels, respectively. Together, these results suggest that the taxonomy covers most failures people notice without being prompted by it, while the automatic labels reproduce human judgments closely enough to support large-scale analysis.

\paragraph{Failure Distribution.}
\begin{figure}[!htbp]
  \centering
  \includegraphics[width=\linewidth]{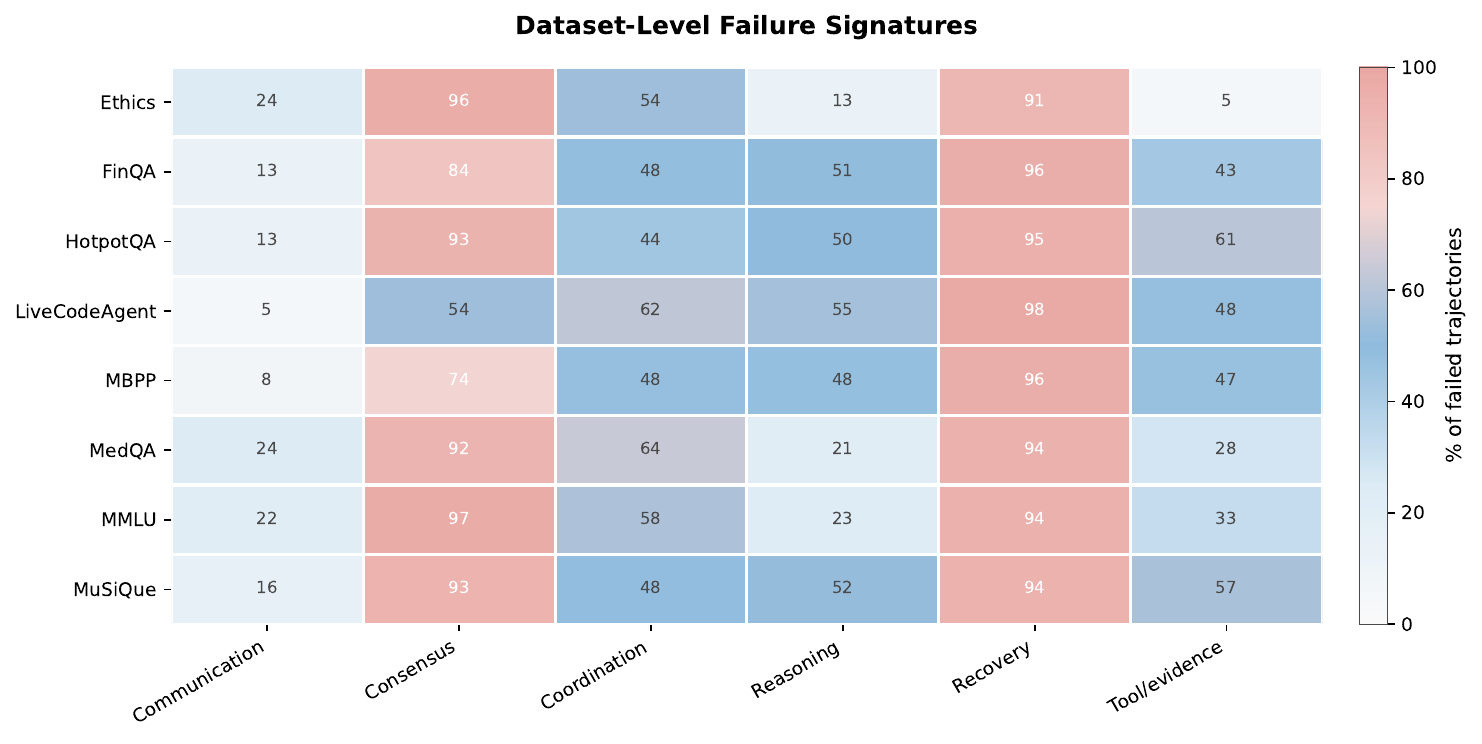}
  \caption{\textbf{Dataset-level failure signatures.} QA-style datasets are more consensus-heavy, while coding, finance, and multi-hop settings expose more reasoning, tool-use, and repair-specific failures.}
  \label{fig:dataset}
\end{figure}

We first examine how failure types vary across datasets. Figure~\ref{fig:dataset} shows that recovery/correction failures are common across all tasks, while consensus failures are especially frequent in QA-style tasks. Coding, finance, and multi-hop tasks show more reasoning and tool/evidence failures, suggesting that the types of errors depend partly on the task. \textbf{The main problem is not always the initial mistake, but the failure to correct it.} Agents may receive corrections, alternative answers, or new evidence during discussion but still fail to update their final answer. This suggests that multi-agent systems need better ways to handle disagreement, verify evidence, and reconsider earlier decisions, rather than simply generating more answers or adding more rounds of debate.

\begin{figure}[!htbp]
  \centering
  \includegraphics[width=\linewidth]{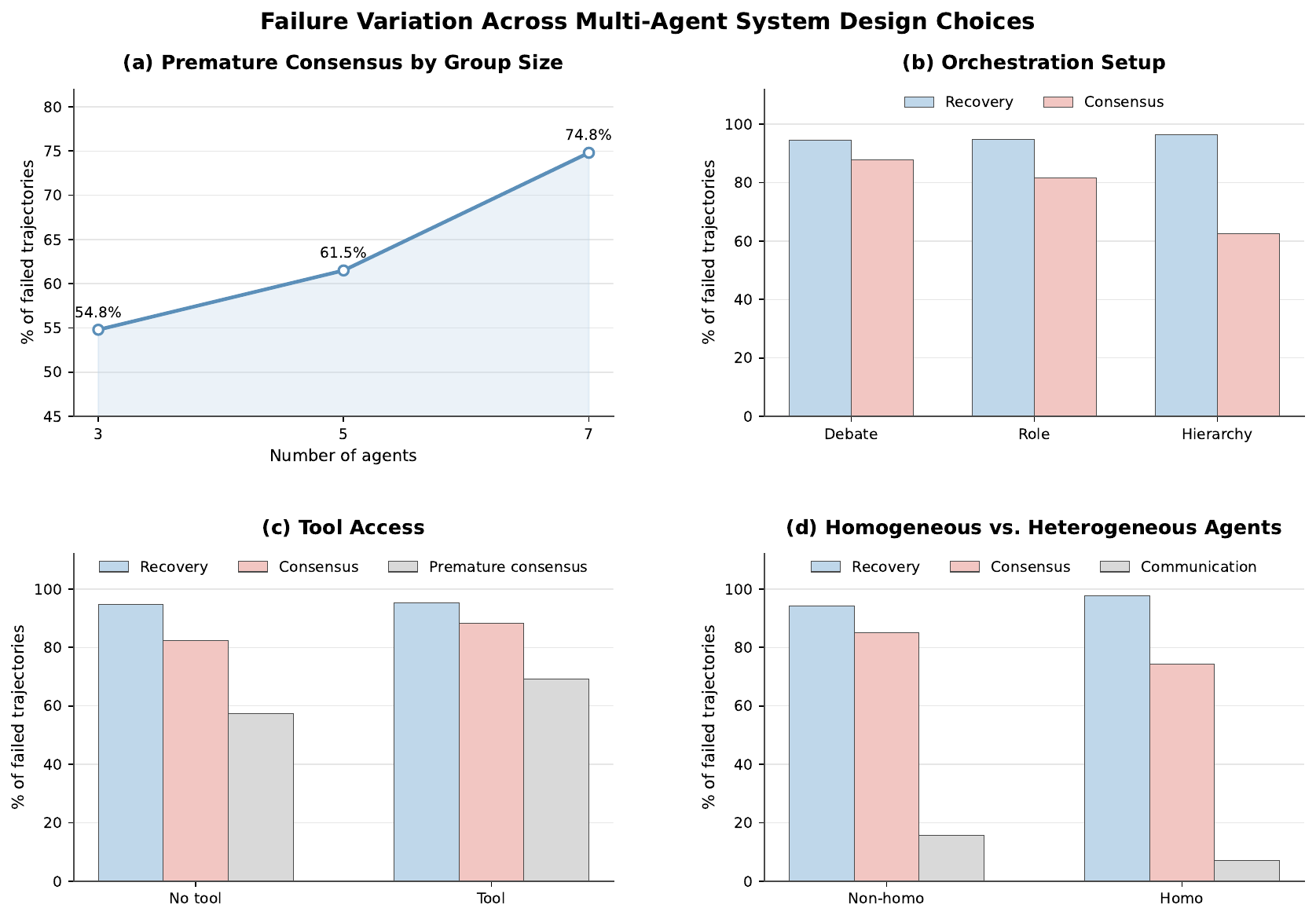}
  \caption{\textbf{Failure variation across system design choices.} Agent count, tool access, orchestration style, and homogeneous composition shift the distribution of failures, but recovery and consensus problems remain common.}
  \label{fig:system_design_effects}
\end{figure}

\paragraph{What drives failures?}
We examine how system-design choices affect failures by comparing configurations that differ in only one factor on the same questions, using question fixed effects, clustered standard errors, and Benjamini--Hochberg correction (App.~\ref{app:design}).

\textbf{Evidence access has the largest effect.} Retrieval reduces the overall failure rate from 74.6\% to 36.4\% on HotpotQA and MuSiQue (95\% CI: $-43.8$ to $-32.7$\,pp). Although tool-enabled systems show a higher proportion of consensus errors among failed debates (Fig.~\ref{fig:system_design_effects}), retrieval reduces their overall occurrence, including premature consensus and wrong framing. In contrast, increasing the team from three to seven agents does not significantly reduce failures. Reasoning errors become less common, but failures after abstention and suppression of correct minority answers increase. Thus, more agents may provide additional reasoning without improving how groups handle uncertainty or disagreement. Hierarchy reduces oscillation but not overall failures, while explicit roles have no detectable effect. Qwen-14B also fails more often than GPT on MMLU, mainly through answer drift (App.~\ref{app:design}; Table~\ref{tab:accuracy_by_agent_configuration}).

Figure~\ref{fig:system_design_effects} also shows that homogeneous teams have more recovery failures but fewer communication failures than heterogeneous teams. This suggests that agents using the same model may coordinate easily but share blind spots, allowing incorrect assumptions to remain unchallenged. Overall, \textbf{better evidence reduces failures, while changes to team structure mainly affect how systems fail.} Premature consensus, wrong framing, and ignored corrections remain statistically unchanged across team size, roles, hierarchy, and model choice (App.~\ref{app:design}). These findings suggest that agent design should prioritize evidence verification, disagreement handling, and correction rather than simply adding more agents or changing their roles.

\paragraph{Failure vs.\ success contrast.}
We compare successful and failed debates to examine whether the same errors occur in both. Even among 500 successful trajectories, 91.2\% contain at least one error under outcome-blind annotation (App.~\ref{app:contrast}). Table~\ref{tab:contrast} shows that reasoning and tool/evidence errors occur at similar rates in both groups. However, failed debates have substantially more recovery, consensus, and coordination failures, with persistent wrong framing showing the largest difference. \textbf{Thus, the key difference is not whether agents make mistakes, but whether they recover from them.}

\begin{table}[t]
\centering\small
\setlength{\tabcolsep}{4pt}
\begin{tabular}{lrrr}
\toprule
 & Succ. & Fail. & Ratio \\
\midrule
Recovery & 0.37 & 0.94 & 2.6$\times$ \\
Consensus & 0.51 & 0.93 & 1.8$\times$ \\
Coordination & 0.14 & 0.53 & 3.6$\times$ \\
Reasoning & 0.35 & 0.35 & 1.0$\times$ \\
Tool / evidence & 0.39 & 0.38 & 1.0$\times$ \\
\midrule
Persistent wrong framing & 0.08 & 0.60 & 7.5$\times$ \\
Premature consensus & 0.20 & 0.71 & 3.5$\times$ \\
\bottomrule
\end{tabular}
\caption{Error prevalence (proportion of trajectories) in successful vs.\ failed debates.}
\label{tab:contrast}
\end{table}

\begin{figure}[!htbp]
  \centering
  \includegraphics[width=\linewidth]{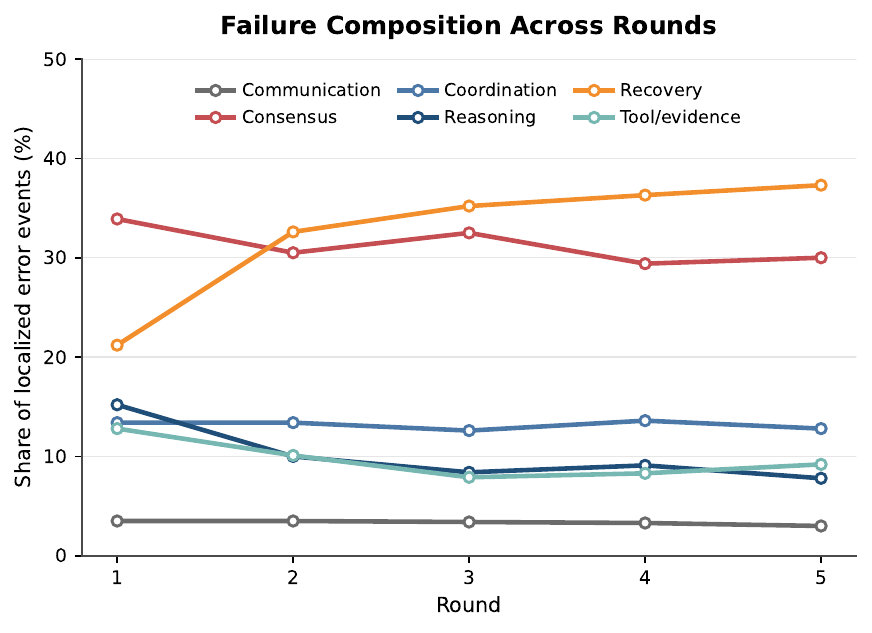}
  \caption{\textbf{Failure composition across deliberation rounds.} Localized error types shift over time: later rounds contain more recovery-oriented failures, while earlier rounds contain more reasoning and consensus-formation errors.}
  \label{fig:round_failure_composition}
\end{figure}

\paragraph{Round-level error dynamics.}
We examine how failure types change across debate rounds. Figure~\ref{fig:round_failure_composition} shows that Round~1 has more individual reasoning and framing errors. In Round~2, after agents see each other's responses, reasoning and evidence related errors decrease and consensus-related errors increase, while later rounds show more recovery failures. \textbf{The first exchange is a critical point where individual mistakes can become group-level failures.} Agents may begin agreeing on an incorrect answer, and further discussion often reinforces it rather than correcting it. This suggests that systems should check evidence and preserve disagreement before agents reach consensus. Later rounds should focus on correcting earlier mistakes rather than simply continuing the debate.

\paragraph{Temporal error cascades.}
We examine whether certain errors are more likely to occur after others, compared with a baseline matched by round and trajectory length (Table~\ref{tab:cascades}). Premature consensus is a common precursor: once agents agree too early, they are more likely to ignore tool results, weak evidence, and later corrections. The reverse also occurs, where ignored evidence and corrections are followed by premature consensus. This suggests that early agreement and failure to use evidence may reinforce each other. Other patterns include edge-case errors followed by failed recalculation and oscillation followed by question misreading, suggesting that unstable discussions may lose focus on the original task.

However, errors that frequently co-occur do not always show a temporal relationship. For example, premature consensus and persistent wrong framing often appear together, but their lift is close to 1. \textbf{These findings suggest that early consensus is an important point for intervention.} Systems could require evidence checks, preserve dissenting answers, or reconsider agreement when new corrections appear. Such interventions may help prevent later errors, although the observed relationships do not establish causality.

\begin{table}[!t]
\centering
\scriptsize
\renewcommand{\arraystretch}{0.95}
\begin{tabular}{@{}l@{\hspace{6pt}}l@{\hspace{2pt}}c@{\hspace{3pt}}c@{}}
\toprule
\textbf{Earlier error} & \textbf{Later error} & 
\textbf{$P(\text{later}\,|\,\text{earlier})$} & \textbf{Lift} \\
\midrule
Premature consensus & Tool result ignored & 11.6\% & 2.79 \\
Premature consensus & Unchallenged evidence & 5.4\% & 1.87 \\
Premature consensus & Correction ignored & 16.5\% & 1.62 \\
Edge-case failure & Failed recalculation & 22.5\% & 1.51 \\
Unsupported assumption & Tool result ignored & 11.4\% & 1.49 \\
Oscillation & Question misread & 10.4\% & 1.39 \\
Tool result ignored & Premature consensus & 70.7\% & 1.37 \\
Premature consensus & Unknown not recovered & 25.5\% & 1.35 \\
Edge-case failure & Unresolved disagreement & 24.3\% & 1.30 \\
Correction ignored & Premature consensus & 60.4\% & 1.22 \\
\bottomrule
\end{tabular}
\caption{\textbf{Cross-type temporal error cascades.} Each row reports how often a later failure type appears after an earlier failure type first appears in the same trajectory. Lift compares this with how often the later error appears after the same round in debates without the earlier error; values above 1 indicate that the later error is unusually likely after the earlier error.}
\label{tab:cascades}
\end{table}

\begin{figure}[!htbp]
  \centering
  \includegraphics[width=\linewidth]{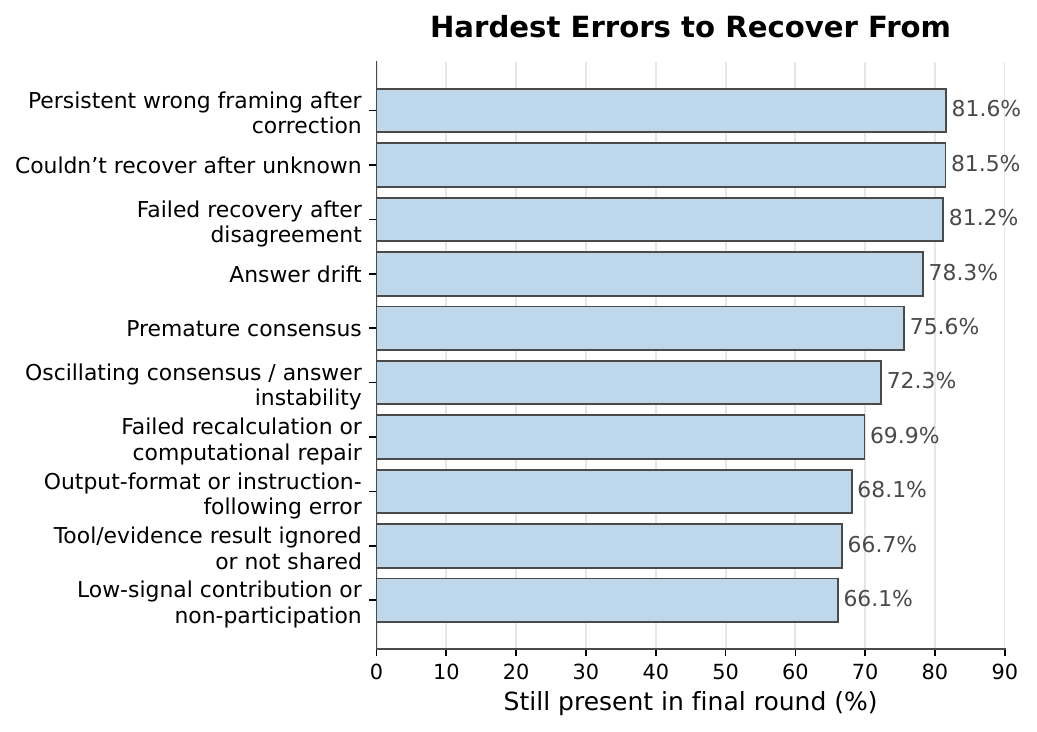}
  \caption{\textbf{Hardest mapped suberrors to recover from.} Bars show the
probability that an error appearing before a debate's final round is still
present in that final round. Group-dynamic and repair-oriented failures are the
most persistent.}
  \label{fig:recovery_hardness_mapped_suberrors}
\end{figure}

\paragraph{Recovery hardness.}
We examine which errors agents can recover from during discussion. Figure~\ref{fig:recovery_hardness_mapped_suberrors} shows that local mistakes, such as interpretation mismatch, question misreading, and unsupported assumptions, are easier to correct. In contrast, persistent wrong framing, failed recovery after unknown answers or disagreement, answer drift, and premature or oscillating consensus often remain unresolved until the final round.

\textbf{More debate rounds do not necessarily lead to better recovery.} Once agents agree on an incorrect interpretation, further discussion often preserves it rather than correcting it. This suggests that systems need explicit recovery mechanisms, such as independently reconsidering the question, rechecking evidence, or using a separate agent to challenge the current answer instead of continuing the same discussion.

\paragraph{Initial-majority transitions.}
We examine how the initial majority influences the final answer, including whether wrong majorities recover and whether correct minority agents can change the outcome (Table~\ref{tab:dataset_transition_dynamics}). Across datasets, wrong majorities are more likely to remain wrong than recover, while correct majorities are usually preserved. Correct minority agents are also suppressed more often than they successfully change the group's answer. These patterns vary by task: MedQA (which has more evidence based discussions) shows stronger recovery, while Ethics has more corruption and minority suppression. Correct answers also have shorter explanations in MedQA, MuSiQue, and Ethics, suggesting that longer discussions do not necessarily improve outcomes. \textbf{Debate tends to preserve the majority rather than recognize correct minority answers.} This suggests that agent systems should evaluate evidence rather than simply follow agreement, for example by verifying minority answers, checking their supporting evidence, and resolving disagreements before making the final decision.

\begin{table}[!t]
\centering
\scriptsize
\setlength{\tabcolsep}{4pt}
\renewcommand{\arraystretch}{1.05}
\begin{tabular}{@{}lrrrrrr@{}}
\toprule
\textbf{Dataset} &
\makecell{\textbf{Rec.}} &
\makecell{\textbf{Corrupted}} &
\makecell{\textbf{Failed}\\\textbf{Recovery}} &
\makecell{\textbf{Preserved}} &
\makecell{\textbf{Ind.}\\\textbf{Rescue}} &
\makecell{\textbf{Ind.}\\\textbf{Supp.}} \\
\midrule
FinQA    & 12.83 & 2.00  & 50.33 & 34.83 & 6.17 & 16.33 \\
HotpotQA & 24.92 & 4.83  & 33.83 & 36.42 & 9.92 & 12.33 \\
MMLU     & 13.21 & 1.42  & 32.54 & 52.83 & 7.00 & 9.17  \\
MedQA    & 56.19 & 0.38  & 36.61 & 6.82  & 0.38 & 1.77  \\
MuSiQue  & 16.25 & 1.75  & 62.67 & 19.33 & 6.42 & 15.17 \\
Ethics   & 3.25  & 12.25 & 26.25 & 58.25 & 1.88 & 28.75 \\
\midrule
\textbf{Overall} & \textbf{20.73} & \textbf{3.62} & \textbf{40.60} & \textbf{35.04} & \textbf{5.78} & \textbf{13.50} \\
\bottomrule
\end{tabular}
\caption{\textbf{Initial-majority transition dynamics by dataset.} Recovered, corrupted, failed recovery, and preserved form the main transition partition. Individual rescue and individual suppression are overlapping diagnostics for cases where initially correct minority agents do or do not change the final answer.}
\label{tab:dataset_transition_dynamics}
\end{table}

\paragraph{Real-time trigger models.}
Finally, we test whether the taxonomy can predict errors before they occur, rather than only identify them afterward. Using error types and their onset rounds, we train models to predict failures one round in advance. Table~\ref{tab:trigger} shows that our model achieves the highest precision, accuracy, and AUROC across all three errors. Simple rules achieve high recall by warning on nearly every debate, while the LLM judge performs close to chance for premature consensus and failed recovery. Our model also improves warning accuracy by up to 20 percentage points over never triggering (Table~\ref{tab:trigger-gain}). \textbf{These results show that the taxonomy can support early failure detection, not just analysis.} Different warnings could trigger different actions: verifying evidence before premature consensus, resolving disagreement before failed recovery, or re-querying agents after an unknown answer. This suggests that targeted interventions based on the predicted error may be more useful than applying the same debate process to every trajectory.

\begin{table}[t]
\centering
\small
\setlength{\tabcolsep}{3.5pt}
\begin{tabular}{lcccccc}
\toprule
Metric & \makecell{Ans.\\change} & Split & \makecell{Round\\$\geq$2} & \makecell{Round\\+split} & \makecell{LLM\\judge} & Ours \\
\midrule
\multicolumn{7}{l}{\textit{Premature consensus}} \\
Prec.  & 0.23 & 0.45 & 0.22 & 0.45 & 0.46 & \textbf{0.60} \\
Rec.   & 0.30 & \textbf{0.97} & 0.34 & \textbf{0.97} & 0.46 & 0.69 \\
F1     & 0.26 & 0.61 & 0.26 & 0.62 & 0.46 & \textbf{0.64} \\
Acc.   & 0.24 & 0.46 & 0.17 & 0.47 & 0.53 & \textbf{0.66} \\
AUROC  & 0.39 & 0.56 & 0.33 & 0.70 & 0.57 & \textbf{0.75} \\
\midrule
\multicolumn{7}{l}{\textit{Failed recovery after disagreement}} \\
Prec.  & 0.16 & 0.46 & 0.15 & 0.45 & 0.46 & \textbf{0.52} \\
Rec.   & 0.19 & \textbf{0.96} & 0.21 & 0.94 & 0.33 & 0.70 \\
F1     & 0.18 & \textbf{0.62} & 0.17 & 0.61 & 0.38 & 0.60 \\
Acc.   & 0.19 & 0.47 & 0.12 & 0.47 & 0.52 & \textbf{0.58} \\
AUROC  & 0.35 & 0.57 & 0.27 & 0.72 & 0.58 & \textbf{0.78} \\
\midrule
\multicolumn{7}{l}{\textit{Couldn't recover after unknown}} \\
Prec.  & 0.09 & 0.32 & 0.09 & 0.37 & 0.72 & \textbf{0.88} \\
Rec.   & 0.19 & \textbf{0.98} & 0.21 & 0.89 & 0.63 & 0.74 \\
F1     & 0.12 & 0.48 & 0.12 & 0.53 & 0.67 & \textbf{0.81} \\
Acc.   & 0.18 & 0.34 & 0.09 & 0.51 & 0.81 & \textbf{0.89} \\
AUROC  & 0.32 & 0.56 & 0.26 & 0.78 & 0.79 & \textbf{0.96} \\
\bottomrule
\end{tabular}
\caption{\textbf{Predicting errors one round before they start.} After each round, each method predicts whether the error begins in the next round. Precision, recall, F1, and accuracy are computed per debate: a warning counts only if it fires before onset. AUROC is computed per round. Parentheses give affected debates over all debates; correct and incorrect debates are balanced. Leave-one-pair-out evaluation; best value per row in bold.}
\label{tab:trigger}
\end{table}

\begin{table}[t]
\centering
\small
\begin{tabular}{lccc}
\toprule
Error & No trigger & Ours & $\Delta$ \\
\midrule
Premature consensus & 0.56 & 0.66 & +0.10 \\
Failed recovery     & 0.55 & 0.58 & +0.03 \\
Unknown recovery    & 0.69 & 0.89 & +0.20 \\
\bottomrule
\end{tabular}
\caption{\textbf{Accuracy with and without the trigger.} Debate-level accuracy when never warning versus with our trigger.}
\label{tab:trigger-gain}
\end{table}

%% file: Sections/06Suggestions.tex
\section{Recommendations for Multi-Agent System Design}

Based on our findings, we highlight four design recommendations.

\begin{itemize}[leftmargin=0pt,itemsep=1pt,topsep=2pt,parsep=0pt,partopsep=0pt]

\item \textbf{Do not rely on self-reported confidence.}
Agents become more confident even when their answers are wrong (Table~\ref{tab:confidence_summary}). Confidence should therefore not be used alone to determine which agent to trust or when to stop debating.

\item \textbf{Use active supervision rather than simple majority voting.}
Supervisors should question unsupported claims, track disagreements, and verify competing answers rather than simply follow the majority. Hierarchy alone does not reliably improve accuracy.

\item \textbf{Use task-specific verification.}
Different tasks require different checks: numerical tasks need calculation verification, coding tasks need edge-case testing, multi-hop QA needs evidence checking, and ethics-style tasks need stronger protection of correct minority answers.

\item \textbf{Intervene based on the error type.}
Premature consensus requires evidence checks, wrong framing requires reconsidering the question, and oscillation requires resolving conflicting answers. Early-warning triggers can help identify when such interventions may be needed, rather than simply adding more debate rounds.

\end{itemize}

\section{Conclusion}

We present a corpus-grounded taxonomy of communication failures in multi-agent LLM systems. Our analysis shows that failures arise not only from incorrect reasoning but also from premature consensus, ignored corrections, and unsuccessful recovery. These findings highlight the importance of verifying evidence, preserving disagreement, and correcting errors during deliberation. The taxonomy also supports early failure detection, providing a basis for more targeted monitoring and intervention in multi-agent systems.

%% file: Sections/Limitations.tex
\section*{Limitations}

Our study focuses on incorrect multi-round trajectories because our goal is to analyze communication failures that emerge through deliberation. This design is useful for studying failure propagation, recovery, and consensus behavior, but it does not capture all possible issues in correct trajectories or in single-turn multi-agent settings. We partially address this by using correct trajectories as a contrast set in analyses such as majority-transition dynamics and trigger-model evaluation.

Our experiments also do not cover the full space of real-world multi-agent environments. We focus on communication-centered deliberation, where agents exchange answers, reasoning, evidence, disagreement, and revisions across rounds. As a result, our corpus does not fully capture GAIA-style tool-heavy tasks, software-development agents, shared-blackboard collaboration, or long-horizon environments where agents interact mainly through files, memory, tools, or external state. These settings may surface additional failures around planning, execution, tool chaining, state tracking, persistent memory, and file-level coordination. We therefore view our taxonomy as a communication-focused foundation that should be extended and validated in more open-ended deployed agent environments.

The taxonomy is also constructed from the datasets, models, and orchestration styles included in our corpus. Although the corpus spans several domains and system designs, additional domains may reveal new fine-grained failure types. To reduce overfitting, we use broad parent categories, preserve evidence links for each label, and report taxonomy saturation results showing that high-level categories stabilize relatively quickly.

Finally, some stages of taxonomy construction and evaluation use LLMs. This may introduce model-specific bias in label generation, mapping, or semantic evaluation. We mitigate this by using structured trajectory cards, explicit unmapped options, iterative clustering, evidence-backed labels, and separate structural checks for granularity, sibling coherence, alignment, relevance, and coverage.

\paragraph{Potential risks.}
This work analyzes failures in multi-agent LLM systems and could be misused to make unreliable systems appear more trustworthy by adding superficial monitoring without meaningful validation. To avoid this, we frame the taxonomy as a diagnostic tool rather than a guarantee of correctness. Our recommendations emphasize verification, disagreement preservation, evidence checking, and human review for high-risk settings. Another risk is that automated failure labels may be treated as definitive explanations of why a system failed. In practice, multi-agent failures can involve several interacting causes, and our taxonomy is multi-label for this reason. We therefore recommend using the labels as structured evidence for analysis and debugging, not as single ground-truth causal explanations.

\paragraph{Reproducibility.}
We report the datasets, model families, orchestration settings, sampling procedure, inference temperatures, and evaluation protocol used in our experiments. Each experiment variant uses a fixed sample where available and seed 42 for sampling. We also log per-round agent answers, confidence scores, reasoning, consensus decisions, moderator or supervisor outputs, final answers, and correctness labels. To support reproducibility, we describe the taxonomy-construction pipeline, including trajectory-card generation, candidate-tag extraction, broad-category mapping, deduplication, unmapped-label clustering, width expansion, depth expansion, and final remapping. We also provide prompt templates and trigger-model details, including features, splits, thresholds, and hyperparameters. Some exact costs may vary across providers and time, but we report approximate token usage and inference costs to make the scale of the experiments transparent.

\section*{Ethical Considerations}

Our work studies communication failures in multi-agent LLM systems with the goal of improving transparency and reliability. The datasets used are existing benchmark datasets, and our analysis focuses on system behavior rather than human subjects. We do not use private user data or attempt to infer sensitive personal attributes. Because multi-agent systems may be used in sensitive domains, including medical, legal, financial, or educational settings, our findings should not be interpreted as showing that debate-based systems are inherently reliable. Instead, the results suggest that multi-agent systems require stronger safeguards, especially around premature consensus, ignored corrections, unsupported confidence, and failed recovery. For high-stakes use cases, taxonomy-based monitoring should be combined with external validation and human oversight.

\paragraph{AI assistance in writing.}
AI writing tools were used to assist with editing, wording refinement, and organization of the manuscript. All technical claims, experimental results, analyses, and final writing decisions were reviewed and controlled by the authors. AI assistance was not used as a substitute for conducting experiments, validating results, or determining the paper's scientific conclusions.

%% file: Sections/Appendix.tex
\section{Experimental Setting}
\label{sec:experimental_setting}

\paragraph{Model Selection and Evaluation Scale.}
We evaluated multi-agent LLM systems across eight datasets spanning financial QA, multi-hop QA, medical QA, ethical judgment, multiple-choice reasoning, and code generation: FinQA, HotpotQA, MBPP, MMLU/MMLU-Pro, MedQA, MuSiQue, Ethics, and LiveCodeBench. Our open-weight agents included LLaMA-3.2-3B, Mistral-7B, Qwen-14B, LLaMA-70B, GPT-OSS-20B, GPT-OSS-120B, and Gemma-12B, with model assignments varying by dataset and configuration. Open-weight inference was run through a combination of GPU infrastructure and DeepInfra-hosted endpoints, while OpenAI endpoints were used for selected taxonomy-construction, mapping, cleanup, and judge-style classification components. Across all experiment files, we ran 7,300 task instances, of which 3,543 ended incorrectly. Since our focus is on failures that emerge through interaction rather than isolated first-turn errors, we filtered to incorrect trajectories with more than one debate round, yielding 3,211 traces for taxonomy construction: 266 from FinQA, 404 from HotpotQA, 320 from MBPP, 325 from MMLU/MMLU-Pro, 281 from MedQA, 755 from MuSiQue, 246 from Ethics, and 614 from LiveCodeBench. For each experiment variant, we used a fixed 100-example sample where available and sampled with seed 42 for reproducibility.

\paragraph{Inference Configuration.}
Most multi-agent debate experiments used temperature \(=0.7\), allowing agents to produce diverse reasoning while keeping the overall protocol consistent. We used lower or role-specific temperatures in settings where the agent role required more controlled behavior. In FinQA, tool and role agents used temperatures between 0.4 and 0.6: normal tool agents generally used 0.5, evidence and calculator agents used 0.4, and skeptic agents used 0.6. In MBPP and LiveCodeAgent homogeneous setups, temperatures varied by role, typically from 0.3 to 0.8; writer agents used higher temperatures around 0.7--0.8, while reviewers, editors, and finalizers used lower temperatures around 0.3--0.5. Moderators and supervisors were run more deterministically, generally at temperature 0.0 or 0.2. Taxonomy benchmark and judge-style classification scripts used temperature 0.0.

Most non-code tasks used a round-based debate protocol with at most five rounds. In round one, agents answered independently with a final answer and confidence; in later rounds, they saw other agents' previous answers, confidence scores, and explanations before revising. Standard debates ended at consensus or the round limit; if no consensus was reached, a non-intervening moderator selected the final answer using only agents' final answers and explanations. In hierarchical runs, a supervisor actively reviewed responses, asked clarifying or follow-up questions, provided guidance, and made the final decision when needed. We varied interaction design across question-only, tool-enabled, role-based, hierarchical, and homogeneous code settings, typically with 3, 5, or 7 agents. Role-based runs included agents such as EvidenceReader, Skeptic, Verifier, Synthesizer, Retriever, Comparator, Deontologist, and Consequentialist.

Dataset-specific pipelines preserved task constraints: FinQA allowed calculator calls through explicit CALC: lines; HotpotQA and MuSiQue included evidence/context variants; MedQA used textbook retrieval and option-letter evaluation; MMLU/MMLU-Pro outputs were normalized to one choice letter; Ethics outputs to binary labels; and MBPP/LiveCodeBench used multi-round code generation while hidden tests, errors, and gold code remained unseen until final evaluation. For each run, we logged the prompt, gold answer, final answer, correctness, rounds to consensus, moderator or supervisor actions, and each agent's per-round answer, confidence, reasoning, and raw response.

\paragraph{Hardware Infrastructure and Cost.}
ARC-hosted open-weight model evaluations were conducted on a GPU cluster with NVIDIA A100 80GB GPUs. Smaller models were run on single GPUs, while larger models required 2--4 GPUs with tensor parallelism. Additional open-weight debate-generation runs were served through DeepInfra endpoints, which cost approximately \$25 in observed billing. Because provider-side billing did not expose uniform token accounting for every model and run, we estimate token usage from logged prompts, responses, number of agents, and number of rounds. The full trajectory-generation stage used approximately 25--35M input tokens and 8--12M output tokens. The taxonomy-construction stage was more input-heavy: each failed multi-round trace was converted into trajectory cards and then processed through candidate-tag extraction, taxonomy mapping, unmapped-label clustering, width/depth expansion, and cleanup. Across the 3,211 taxonomy traces, this required approximately 25--37M additional input tokens and 6--10M output tokens through OpenAI endpoints. Overall, the full experimental and taxonomy-construction pipeline cost approximately \$38--50 in API inference, with most of the direct API cost coming from multi-agent debate generation. Additional details are in Table \ref{tab:cost}.

\begin{table*}[!t]
\centering
\scriptsize
\setlength{\tabcolsep}{4pt}
\renewcommand{\arraystretch}{1.08}
\begin{tabular}{@{}lrrrr@{}}
\toprule
\textbf{Component} &
\textbf{Provider / Infrastructure} &
\textbf{Approx. input toks.} &
\textbf{Approx. output toks.} &
\textbf{Approx. cost} \\
\midrule
Multi-agent trajectory generation &
GPU cluster / DeepInfra &
25--35M &
8--12M &
\$25.00 \\
Trajectory card generation &
OpenAI &
14--18M &
3--5M &
\$6--10 \\
Candidate tag extraction &
OpenAI &
4--6M &
1.5--2.5M &
\$3--5 \\
Taxonomy mapping and cleanup &
OpenAI &
6--10M &
1--2M &
\$3--6 \\
Cluster review and width/depth expansion &
OpenAI &
1--3M &
0.3--0.8M &
\$1--3 \\
LLM-judge trigger baseline &
OpenAI &
$<$0.1M &
$<$0.1M &
$<$\$1 \\
\midrule
\textbf{Total} &
-- &
\textbf{50--72M} &
\textbf{14--22M} &
\textbf{\$38--50} \\
\bottomrule
\end{tabular}
\caption{Approximate inference cost by pipeline component. DeepInfra cost reflects observed billing. ARC runs used institutional GPU infrastructure and are therefore reported as compute usage rather than direct API cost. OpenAI costs are estimated from approximate input/output token counts and provider-reported per-token prices at the time of experimentation.}
\label{tab:cost}
\end{table*}

\[
\text{Cost} =
\frac{T_{\text{in}}}{10^6} C_{\text{in}}
+
\frac{T_{\text{out}}}{10^6} C_{\text{out}},
\]
where \(T_{\text{in}}\) and \(T_{\text{out}}\) are approximate input and output tokens, and \(C_{\text{in}}\) and \(C_{\text{out}}\) are the provider-specific prices per million tokens.

\section{Expanded Related Work}
Table \ref{tab:taxonomy_dataset_comparison} and \ref{tab:related_empirical_diagnostic} distinguish our work from related works.

\begin{table*}[t]
\centering
\scriptsize
\setlength{\tabcolsep}{3.2pt}
\renewcommand{\arraystretch}{1.12}
\resizebox{\textwidth}{!}{
\begin{tabular}{p{2.9cm} c c c c p{5.0cm} c c}
\toprule
\textbf{Work} &
\textbf{Comm. Error?} &
\textbf{Automatic?} &
\textbf{\# Domains} &
\textbf{\# Traces} &
\textbf{Failure Categories / Focus} &
\textbf{Tracks Answer Evolution?} &
\textbf{Multi-error?} \\
\midrule

\cite{cemri2026multi} &
\pmark &
\xmark &
7 frameworks &
1600+ &
System design; inter-agent misalignment; task verification &
\xmark &
\cmark \\

\cite{zhang2025agent} &
\xmark &
\pmark &
2 datasets &
127 systems / 184 annotations &
Responsible-agent identification; decisive failure-step attribution &
\pmark &
\xmark \\

\cite{zhu2025llm}  &
\xmark &
\pmark &
3 tasks &
\nr &
Planning; tool use; memory; reflection; action; system-level failures &
\xmark &
\cmark \\

\cite{bi2026diagnosing} &
\xmark &
\xmark &
10 systems &
307 &
Root causes in platform-orchestrated agentic workflows; platform/system failures &
\xmark &
\cmark \\

\cite{barke2026agentrx} &
\xmark &
\cmark &
3 domains &
115 &
Critical failure steps; root-cause attribution; repair-oriented diagnosis &
\xmark &
\cmark \\

\cite{ma2026demystifying} &
\xmark &
\xmark &
2 platforms / 10 systems &
307 &
Lifecycle failures; planning; execution; tool use; orchestration failures &
\xmark &
\cmark \\

\cite{lu2025exploring} &
\xmark &
\xmark &
34 tasks &
204 runs / 104 failures &
Planning; execution; response generation; tool-related errors &
\xmark &
\cmark \\

\cite{kong2025aegis} &
\xmark &
\cmark &
6 task domains &
9,533 &
Faulty agents; injected error modes; attribution-oriented failures &
\xmark &
\cmark \\

\cite{in2026rethinking} &
\xmark &
\pmark &
121 MAS configurations &
289 &
Multi-perspective failure attribution; limitations of single-root-cause diagnosis &
\pmark &
\cmark \\

\cite{yu2025correct} &
\xmark &
\cmark &
7 datasets &
2000+ &
Step-level errors; error recognition; correction-related failures &
\xmark &
\cmark \\

\cite{deshpande2025trail} &
\xmark &
\xmark &
\nr &
148 traces / 841 errors &
Reasoning; planning/coordination; system execution failures &
\xmark &
\cmark \\

\cite{chen2026seeing} &
\xmark &
\xmark &
\nr &
\nr &
Multi-perspective failure attribution; responsible component or step &
\pmark &
\xmark \\

\midrule

\textbf{Ours} &
\cmark &
\cmark &
\textbf{10} &
\textbf{7000+// 3000+ errors} &
\textbf{Reasoning; communication; evidence/tool use; coordination; consensus; recovery} &
\cmark &
\cmark \\

\bottomrule
\end{tabular}
}
\caption{
Comparison with prior taxonomy, dataset, and benchmark resources for agent and multi-agent failure analysis.
\textit{Comm. Error?} indicates whether inter-agent communication failures are a primary object of analysis.
\textit{Automatic?} indicates whether taxonomy construction or annotation is automatic/semi-automatic rather than primarily human-authored.
\textit{Tracks Answer Evolution?} indicates whether the work explicitly follows answer changes across agents and rounds.
\textit{Multi-error?} indicates whether multiple failure mechanisms can be assigned to the same trace.
\cmark = yes, \xmark = no, $\triangle$ = partial, \textsc{NR} = not reported.
}
\label{tab:taxonomy_dataset_comparison}
\end{table*}

\begin{table*}[t]
\centering
\scriptsize
\setlength{\tabcolsep}{4pt}
\renewcommand{\arraystretch}{1.15}
\resizebox{\textwidth}{!}{
\begin{tabular}{p{3.0cm} p{2.4cm} c c p{4.2cm} p{6.0cm}}
\toprule
\textbf{Work} &
\textbf{Paper Type} &
\textbf{Comm. Relevant?} &
\textbf{Taxonomy Resource?} &
\textbf{Main Contribution} &
\textbf{Distinction from Our Work} \\
\midrule

\cite{wynn2025talk} &
Empirical debate study &
\cmark &
\xmark &
Shows that multi-agent debate can reduce accuracy through peer influence, agreement pressure, and correctness shifts over rounds. &
This work motivates the importance of communication failures, but it does not construct a general taxonomy or annotation framework for such failures. Our work turns this kind of observation into a broader framework that identifies where communication failures originate, how they propagate, whether they are corrected, and how they affect finalization across tasks and orchestration settings. \\

\cite{bi2026diagnosing} &
Diagnostic method &
\xmark &
\xmark &
Uses behavioral abstractions or diagnostic insights to explain and localize agent failures. &
This work focuses on improving diagnosis through higher-level behavioral summaries, but it does not model how agents influence one another over multiple deliberation rounds. Our work explicitly tracks answer movement, correction attempts, evidence uptake, confidence shifts, dissent, and finalization decisions, making the communication process itself the object of analysis. \\

\bottomrule
\end{tabular}
}
\caption{
Related empirical and diagnostic studies that motivate our work but are not primarily taxonomy, dataset, or benchmark resources for communication-failure analysis.
}
\label{tab:related_empirical_diagnostic}
\end{table*}


\section{Contrast Set Details}\label{app:contrast}
\paragraph{Setup.} We sampled 500 correct multi-round trajectories (83--84 per dataset) from FinQA, HotpotQA, MMLU, MedQA, MuSiQue, and Ethics. Because the card prompt presupposes failure, GPT-5.4-nano instead lists any errors supported by quoted trace evidence (empty lists allowed) without seeing the outcome or reference answer; each error is then mapped to one taxonomy label or left unmapped (0.6\%). Successful trajectories average 2.4 errors. Failure rates use the 2,277 failed trajectories from the same datasets, weighted to the contrast set's dataset mix.

\paragraph{Additional results.} The pattern holds in all six datasets. Communication errors are more frequent in successes (34.6\% vs.\ 18.4\%), mostly minor issues such as weakly challenged reasoning that the outcome-blind prompt reports but the failure-focused pipeline did not; we therefore interpret only large gaps. Moderator/hierarchy failure (12.0\% vs.\ 42.5\%) and oscillating consensus (14.4\% vs.\ 28.7\%) are also more frequent in failures.

\section{Design-Factor Analysis}
\label{app:design}
We match taxonomy cards to their source debates by question number and verify every match against the gold answer; all 2,277 cards from the six non-coding datasets match. Coding cards (MBPP, LiveCodeBench) cannot be verified this way, so coding datasets are used only for failure rates.
For each design factor, we pair configurations that differ only in that factor (e.g., HotpotQA tools vs.\ question-only with 3, 5, and 7 agents) and keep questions run in both.
We fit linear probability models with pair$\times$question fixed effects and standard errors clustered by question, and apply Benjamini--Hochberg correction across all 252 tests.
For error types, the outcome is a failed debate tagged with that error. Failed debates without a taxonomy card (coverage 51--100\% per configuration) are dropped, and carded failures are reweighted by their configuration's coverage.
\emph{Share} outcomes condition on failure, so we read them as changes in the failure mix.
Because we assigned the configurations and the questions are fixed, the estimates measure the effect of each design choice; they do not cover factors we did not vary.
Heterogeneous and homogeneous teams were run on different question sets, so we do not compare them.

\section{Trigger Model Details}
\label{app:trigger}

\paragraph{Data.} The trigger set contains 304 debates from eight datasets: 152 question-matched pairs, each with one correct and one incorrect debate on the same question, and 19 pairs per dataset. We prefer debates with at least three rounds; 8 debates have two rounds. Correctness is used only to build the balanced sample and is never shown to the annotator or the models.

\paragraph{Labels.} Each debate is annotated once by an outcome-blind LLM annotator (\texttt{gpt-6-luna}) for three labels. The annotator sees each agent's per-round responses, without reference answers, correctness, or duplicated extracted-answer fields. For each label it returns present, absent, or uncertain, and for each occurrence the round, the agents involved, and verbatim evidence quotes, which are validated against the transcript. Premature consensus uses the taxonomy definition with explicit minimum conditions. The taxonomy definition of failed recovery after disagreement refers to the correct final answer, so we use an outcome-blind version: a visible disagreement is never adjudicated on substance, and the debate either ends split or is settled by majority, vote, confidence, authority, or one side going quiet. Couldn't recover after unknown requires an abstention before the last round, followed by a later round in which the agent abstains again or copies another answer without engaging its evidence. The onset is the earliest occurrence round. Uncertain labels are excluded for that error. Labeled prevalence is 134/304 (premature consensus), 136/303 (failed recovery), and 92/299 (unknown recovery), split nearly evenly between correct and incorrect debates. To estimate label consistency, premature consensus was annotated in two independent runs with near-identical definitions: they agree on 81.2\% of debates ($\kappa = 0.62$), and on the onset round in 80.6\% of debates both runs mark present.

\paragraph{Prediction task.} A row is the state of a debate after round $t$, for every $t$ before the last round and before the error's onset. Its label is 1 if the onset is round $t{+}1$ and 0 otherwise; debates without the error contribute only negative rows. This gives 767, 705, and 816 rows for the three labels.

\paragraph{Features.} The model uses 40 features: the round index plus answer-dynamics features computed from each agent's extracted answer and stated confidence in rounds $1..t$. Split shape: number of answering agents, distinct answers, majority share, margin over the runner-up, normalized entropy, minority size, lone dissenter, near tie, and unanimity. Change since round $t{-}1$: number and fraction of agents that changed; agents that moved to the current or previous majority, left the majority, or moved to a new answer; majority change; whether the group just became unanimous; and the change in distinct answers, majority share, and entropy. History: flip-backs ($A{\rightarrow}B{\rightarrow}A$) this round and in total, total changes, agents that ever changed, the maximum changes by one agent, agents still holding their round-1 answer, majority changes, answers seen, new answers this round, unanimity streak, and rounds spent split. Confidence: whether confidences are reported, and their mean, minimum, standard deviation, change, the majority--minority gap, the mean confidence of agents who changed, and the count below 80\%. The model uses no message text and no dataset identity.

\paragraph{Model and evaluation.} We use \texttt{RandomForestClassifier} with 300 trees, minimum leaf size 5, \texttt{max\_features="sqrt"}, and balanced class weights. Evaluation is leave-one-pair-out: in each of 152 folds, both debates of one question are held out. The warning threshold is chosen on the training fold as the highest threshold that reaches 80\% training recall. At the debate level, a debate with the error counts as caught if the trigger fires at any round before onset; a debate without the error counts as a false alarm if the trigger fires at any round. AUROC is computed over held-out rows.

\paragraph{Baselines.} \emph{Answer change} fires when any agent's answer differs from its previous-round answer. \emph{Split} fires when agents give more than one distinct answer. \emph{Round $\geq$2} fires from round 2 onward. \emph{Round+split} is a logistic regression on the round index, a split indicator, and the number of distinct answers, trained with the same folds and threshold rule as our model. The \emph{LLM judge} is GPT-4.1-mini. For each debate and round $t$, it receives the task and all messages up to round $t$, outcome-blind and with no later rounds, along with short definitions of the errors, and answers yes or no to whether each error starts in round $t{+}1$ (1{,}017 requests). Because the judge answers yes or no, its AUROC is computed from binary scores. For couldn't recover after unknown, a rule that fires whenever an agent abstains in round $t$ is competitive (precision .74, recall .92), reflecting how visible the precursor is; we report the learned model for consistency across labels.

\section{Additional Failure Distribution Results}
\label{app:failure_distribution}

We report additional descriptive statistics for the failure-distribution analysis. All counts are computed over 3,211 failed multi-agent trajectories. Because our taxonomy is multi-label, a single trajectory can contain multiple broad failures and multiple fine-grained suberrors, so percentages do not sum to 100\%.

Table~\ref{tab:top_suberrors} shows the most frequent fine-grained failures. The dominant mapped suberrors are persistent wrong framing after correction and premature consensus, which appear in 62.1\% and 59.6\% of failed trajectories, respectively. At the child-suberror level, anchor-driven recovery failure, social convergence without verification, and unstable coordination are the most common. This reinforces the main finding that failures are rarely just local mistakes; they often involve agents converging around, preserving, or failing to repair an incorrect frame.

Table~\ref{tab:recovery_cooccurrence} shows that recovery failures frequently co-occur with other broad failure types. Recovery overlaps with consensus failures in 78.6\% of failed trajectories, and also commonly appears with coordination, tool/evidence, and reasoning failures. This suggests that recovery is not a separate late-stage cleanup issue. Instead, it is entangled with how agents interpret disagreement, use evidence, coordinate revisions, and decide when to finalize an answer.

Table~\ref{tab:overrepresented_suberrors} shows which suberrors are disproportionately associated with particular datasets or system settings. FinQA strongly overrepresents calculation and representation errors, while LiveCodeAgent overrepresents edge-case handling and failed computational repair. MBPP shows higher failed recovery after disagreement. In the agent-count setting, 7-agent runs overrepresent both failure to recover after unknown states and low-signal contribution. These patterns show that failure mechanisms are task- and design-dependent: changing the dataset or the number of agents shifts the kind of failure that becomes most likely.

\begin{figure}[!htbp]
  \centering
  \includegraphics[width=\linewidth]{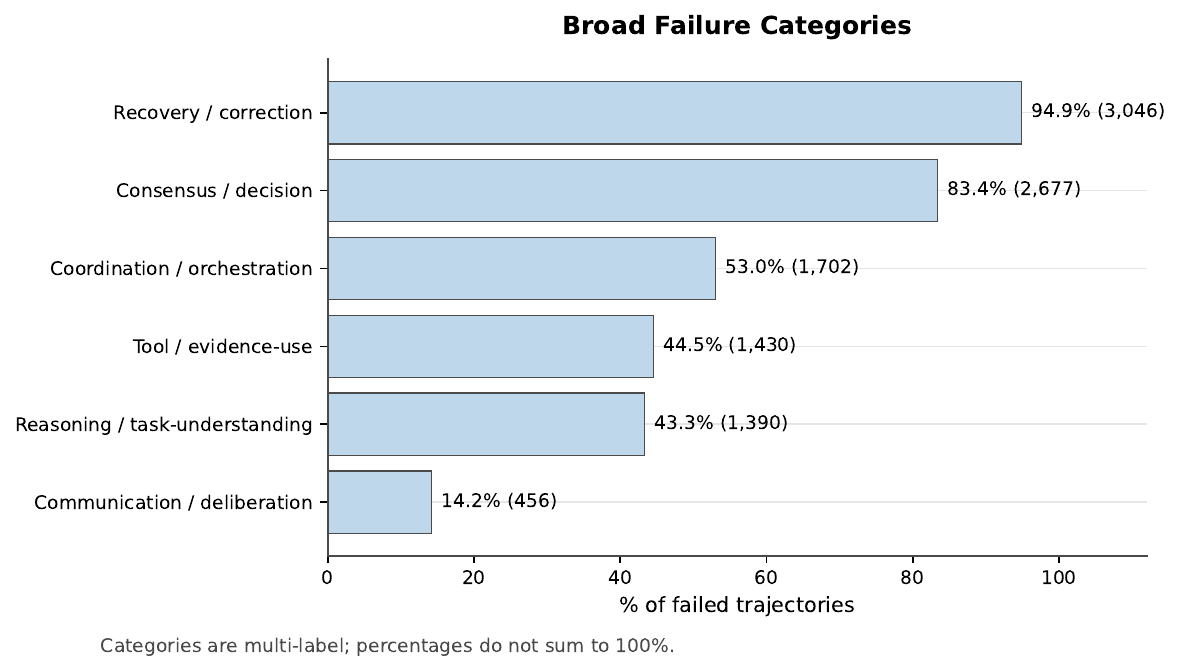}
  \caption{\textbf{Broad failure category distribution.} Recovery/correction and consensus/decision failures dominate the failed trajectory set, indicating that many errors are failures of deliberative repair rather than isolated first-pass mistakes.}
  \label{fig:broad_failure_distribution}
\end{figure}

Figure~\ref{fig:broad_failure_distribution} reports the broad failure category distribution. Recovery/correction and consensus/decision failures dominate the failed trajectory set, consistent with the fine-grained tables above. Together, these results show that failed multi-agent deliberation is driven less by isolated first-pass errors than by failures of repair, verification, and agreement.

\begin{table*}[!t]
\centering
\scriptsize
\setlength{\tabcolsep}{4pt}
\renewcommand{\arraystretch}{0.95}
\begin{tabular}{@{}llccc@{}}
\toprule
\textbf{Setting} &
\textbf{Overrepresented suberror} &
\textbf{In setting} &
\textbf{Elsewhere} &
\textbf{Lift} \\
\midrule

\rowcolor{gray!10}
\multicolumn{5}{c}{\textbf{\textit{Dataset-specific failures}}} \\
\midrule
FinQA &
Calculation / representation error &
25.9 &
2.7 &
9.55$\times$ \\

LiveCodeAgent &
Edge-case handling failure &
27.9 &
3.0 &
9.27$\times$ \\

LiveCodeAgent &
Failed computational repair &
37.1 &
11.9 &
3.12$\times$ \\

MBPP &
Failed recovery after disagreement &
31.9 &
18.3 &
1.74$\times$ \\

\midrule
\rowcolor{gray!10}
\multicolumn{5}{c}{\textbf{\textit{Agent-count-specific failures}}} \\
\midrule
7 agents &
Could not recover after unknown &
37.3 &
20.9 &
1.78$\times$ \\

7 agents &
Low-signal contribution &
14.7 &
4.4 &
3.34$\times$ \\

\bottomrule
\end{tabular}
\caption{\textbf{Suberrors overrepresented in specific datasets and system settings.}
Percentages report how often a suberror appears within the named setting versus outside it. Lift is the ratio between the two rates.}
\label{tab:overrepresented_suberrors}
\end{table*}

\begin{table}[!t]
\centering
\scriptsize
\setlength{\tabcolsep}{5pt}
\renewcommand{\arraystretch}{0.95}
\begin{tabular}{@{}llcc@{}}
\toprule
\textbf{Level} &
\textbf{Failure subtype} &
\textbf{Count} &
\textbf{\% of cases} \\
\midrule

\rowcolor{gray!10}
\multicolumn{4}{c}{\textbf{\textit{Child suberrors}}} \\
\midrule
Child &
Anchor-driven recovery failure &
1,067 &
33.2 \\

Child &
Social convergence without verification &
1,053 &
32.8 \\

Child &
Unstable coordination &
856 &
26.7 \\

\midrule
\rowcolor{gray!10}
\multicolumn{4}{c}{\textbf{\textit{Mapped suberrors}}} \\
\midrule
Mapped &
Persistent wrong framing after correction &
1,993 &
62.1 \\

Mapped &
Premature consensus &
1,913 &
59.6 \\

Mapped &
Moderator or hierarchy failure &
1,379 &
42.9 \\

\bottomrule
\end{tabular}
\caption{\textbf{Most frequent fine-grained failure subtypes.}
Counts are computed over 3,211 failed multi-agent trajectories. Labels are multi-label, so percentages do not sum to 100\%.}
\label{tab:top_suberrors}
\end{table}

\begin{table}[!t]
\centering
\scriptsize
\setlength{\tabcolsep}{5pt}
\renewcommand{\arraystretch}{0.95}
\begin{tabular}{@{}lcc@{}}
\toprule
\textbf{Failure pair} &
\textbf{Count} &
\textbf{\% of all cases} \\
\midrule

Consensus + Recovery &
2,523 &
78.6 \\

Coordination + Recovery &
1,589 &
49.5 \\

Tool/evidence + Recovery &
1,341 &
41.8 \\

Reasoning + Recovery &
1,318 &
41.0 \\

\bottomrule
\end{tabular}
\caption{\textbf{Co-occurrence between recovery failures and other broad error types.}
Counts are computed over 3,211 failed trajectories. The high overlap shows that many failures involve converting disagreement, evidence, or correction signals into a final repaired answer.}
\label{tab:recovery_cooccurrence}
\end{table}

\section{Additional Round-Level Error Dynamics}
\label{app:round_level_dynamics}

\begin{figure}[!htbp]
  \centering
  \includegraphics[width=\linewidth]{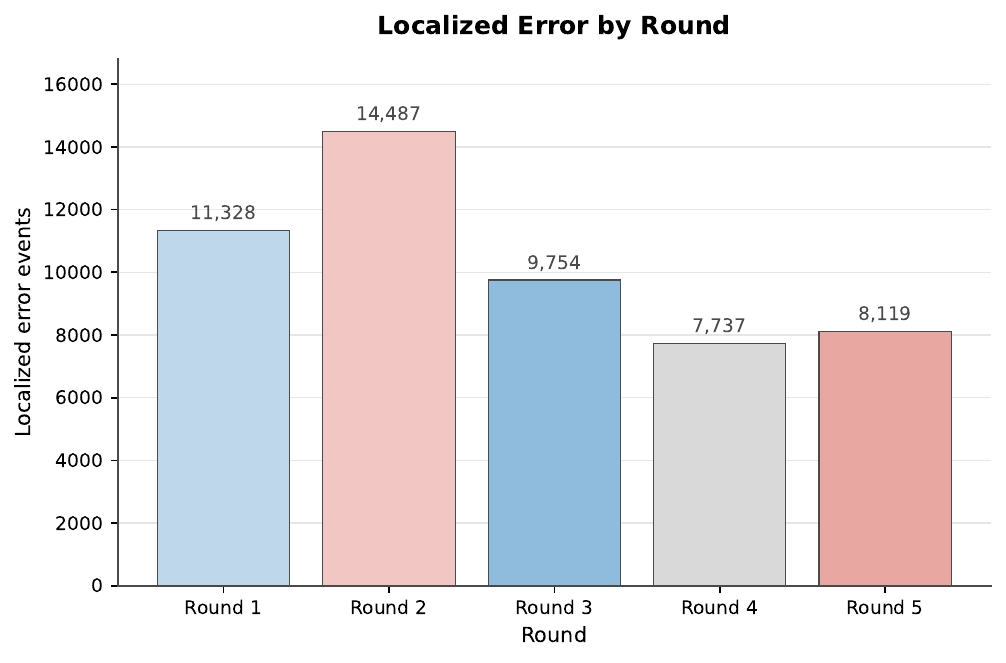}
  \caption{\textbf{Localized error counts by round.} Error events peak after the first exchange, indicating that many failures emerge as agents react to peer responses rather than only in the initial answer.}
  \label{fig:round_localized_error_counts}
\end{figure}

We report additional round-level statistics for the deliberation analysis. Figure~\ref{fig:round_localized_error_counts} shows the number of localized error events by round. Errors peak in Round~2, after agents have seen initial peer responses and begin revising, aligning, or challenging one another. This supports the main result that failures are not only inherited from the first answer. A substantial share emerges during the first interaction step, when agents begin responding to the group context.

The decline in later rounds does not necessarily mean that deliberation has repaired the trajectory. Instead, it is consistent with the round-composition result: later rounds contain fewer new localized errors but a larger share of recovery-oriented failures. In other words, many trajectories enter a bad state after the first exchange and then spend subsequent rounds failing to correct it. This makes Round~2 an important diagnostic point for detecting whether deliberation is becoming corrective or merely stabilizing an early mistake.

\section{Additional Initial-Majority Transition Results}
\label{app:initial_majority_transitions}

We report additional transition results by number of agents. As in the main analysis, \emph{recovered} means the initial majority was wrong but the final answer is correct; \emph{failed recovery} means the initial majority was wrong and remains wrong; \emph{preserved} means the initial majority was correct and stays correct; and \emph{corrupted} means the initial majority was correct but the final answer becomes wrong. We also report two overlapping diagnostics: \emph{individual rescue}, where an initially correct minority signal is associated with a correct final answer despite a wrong initial majority, and \emph{individual suppression}, where an initially correct minority signal is overridden and the final answer remains or becomes wrong.

Table~\ref{tab:agent_transition_dynamics} shows that adding agents does not produce a large improvement in recovery. Recovery increases only modestly from 19.50\% with 3 agents to 22.19\% with 5 agents and 22.06\% with 7 agents, while failed recovery remains near 40\% across all settings. The more consistent effect is on minority signals: individual rescue decreases as group size increases, from 7.87\% with 3 agents to 2.50\% with 7 agents, while individual suppression rises from 11.37\% to 17.70\%. Thus, larger groups do not reliably make deliberation more corrective. They slightly increase recovery but also make it more likely that correct minority answers are absorbed or overridden by the group.

\begin{table}[!t]
\centering
\scriptsize
\setlength{\tabcolsep}{4pt}
\renewcommand{\arraystretch}{1.05}
\begin{tabular}{@{}crrrrrr@{}}
\toprule
\textbf{\#Agents} &
\makecell{\textbf{Recovered}} &
\makecell{\textbf{Corrupted}} &
\makecell{\textbf{Failed}\\\textbf{Recovery}} &
\makecell{\textbf{Preserved}} &
\makecell{\textbf{Ind.}\\\textbf{Rescue}} &
\makecell{\textbf{Ind.}\\\textbf{Supp.}} \\
\midrule
3 & 19.50 & 3.37 & 41.51 & 35.62 & 7.87 & 11.37 \\
5 & 22.19 & 3.78 & 39.27 & 34.76 & 4.11 & 14.67 \\
7 & 22.06 & 4.00 & 40.13 & 33.81 & 2.50 & 17.70 \\
\bottomrule
\end{tabular}
\caption{\textbf{Initial-majority transition dynamics by number of agents.} Recovery rates remain broadly similar as the number of agents increases, while individual suppression rises with group size.}
\label{tab:agent_transition_dynamics}
\end{table}

\section{Model-Only Taxonomy Classification}
\label{app:model_only_taxonomy_classification}

We evaluate whether off-the-shelf LLMs can directly assign our taxonomy labels from whole debate traces. This experiment tests a weaker alternative to our annotation pipeline: instead of using structured trajectory cards, evidence spans, controlled mapping, and taxonomy-guided expansion, we ask models to classify parent and leaf error labels directly.

\begin{figure*}[!htbp]
  \centering
  \includegraphics[width=\linewidth]{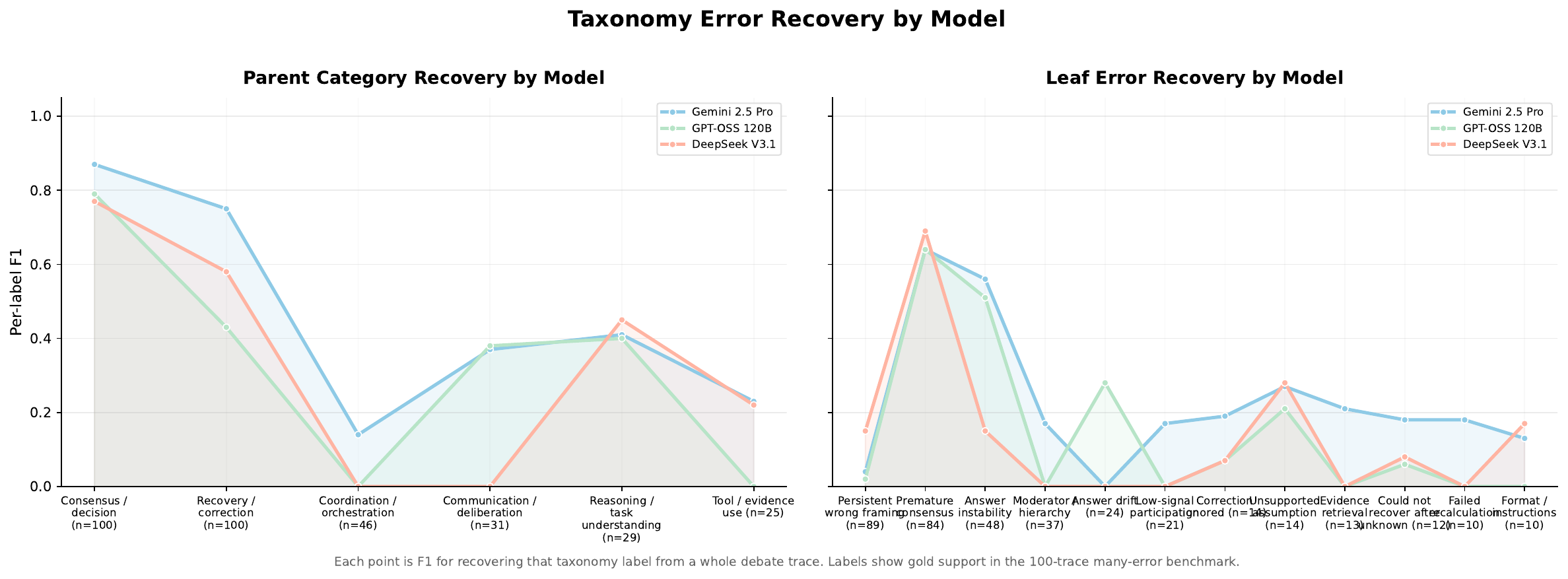}
  \caption{\textbf{Model-only taxonomy classification by error label.} Each point reports per-label F1 when a model classifies parent or leaf error labels from a whole debate trace. Low and uneven performance, especially on fine-grained labels, motivates a systematic taxonomy construction and annotation pipeline.}
  \label{fig:model_taxonomy_classification}
\end{figure*}

Figure~\ref{fig:model_taxonomy_classification} shows that model-only taxonomy classification is not reliable enough for this setting. Performance varies substantially across both models and labels, with especially weak results on fine-grained leaf errors. Models can sometimes recover broad categories such as consensus or recovery failures, but they struggle with labels that require tracking evidence, attributing failures to specific turns, and reasoning about whether a later response actually repaired an earlier mistake. This supports our choice to build the taxonomy through a more structured annotation process rather than relying on unconstrained LLM labeling.

\section{Overall accuracy for all runs}

Table~\ref{tab:accuracy_by_agent_configuration} reports run-level accuracy across datasets, orchestration settings, and agent compositions. Accuracy varies much more by dataset and task setting than by simply adding agents. The strongest results appear on HotpotQA with roles or tools, MMLU, Ethics, and MedQA, where many configurations reach the mid-60s to high-70s. HotpotQA shows the clearest gain from system design: question-only runs are much weaker, while role and tool settings improve accuracy substantially. MuSiQue follows a similar pattern, with very low question-only accuracy but large gains from roles and tools, suggesting that multi-hop tasks benefit from explicit evidence-oriented structure.

Other datasets remain difficult despite multi-agent deliberation. FinQA stays near chance-level performance across tool and tool-role settings, indicating that tool access alone does not solve numerical or representation-heavy reasoning. MBPP is moderate, with most settings in the mid-50s to low-60s, while LiveCodeAgent is consistently low across configurations, showing that program repair and execution-sensitive coding remain especially challenging. MedQA is comparatively stable across settings, with accuracy mostly in the low-to-mid 60s, while MMLU remains relatively strong across homogeneous, debate, and hierarchical variants.

Increasing the number of agents does not produce a uniform accuracy improvement. Some settings improve with larger groups, such as HotpotQA roles and MuSiQue roles, but others are flat or decline, including several MBPP, MMLU, Ethics, and FinQA configurations. This matches the failure analysis: adding agents can provide more evidence or alternative reasoning, but it can also increase consensus pressure, suppression of minority signals, or shared error reinforcement. Overall, the table shows that orchestration and task structure matter more than group size alone, and that multi-agent systems are most useful when the task benefits from explicit evidence aggregation rather than unsupported deliberation.
\newcommand{\agenthead}[2]{%
\begin{tabular}[c]{@{}c@{}}\textbf{#1}\\{\tiny #2}\end{tabular}}

\begin{table*}[!t]
\centering
\scriptsize
\setlength{\tabcolsep}{4pt}
\renewcommand{\arraystretch}{0.95}
\begin{tabular}{@{}lcccc@{}}
\toprule
\textbf{Setting} &
\agenthead{3 Base}{Llama3.2, Mistral, Llama3.3} &
\agenthead{3 Alt}{gpt-oss/Gemma/Qwen} &
\agenthead{5 Agents}{Base + gpt-oss, Gemma} &
\agenthead{7 Agents}{+ Qwen3, gpt-oss-120B} \\
\midrule

\rowcolor{gray!10}
\multicolumn{5}{c}{\textbf{\textit{FinQA}}} \\
\midrule
Tools & 47 & 51 & 48 & -- \\
Tools + Roles & 49 & 47 & 44 & -- \\

\midrule
\rowcolor{gray!10}
\multicolumn{5}{c}{\textbf{\textit{HotpotQA}}} \\
\midrule
Question-only & 39 & 32 & 34 & 37 \\
Roles & 71 & 67 & 75 & 76 \\
Tools & 74 & 78 & 77 & 76 \\

\midrule
\rowcolor{gray!10}
\multicolumn{5}{c}{\textbf{\textit{MBPP}}} \\
\midrule
Hierarchical & 59 & -- & 53 & -- \\
Homogeneous & 54 & 56 & 59 & -- \\
Roles & 54 & 61 & 56 & -- \\

\midrule
\rowcolor{gray!10}
\multicolumn{5}{c}{\textbf{\textit{MMLU}}} \\
\midrule
Homogeneous (GPT) & 71 & -- & 74 & -- \\
Homogeneous (Llama70B) & 61 & -- & 68 & -- \\
Homogeneous (Qwen) & 64 & -- & 61 & -- \\
Debate & 67 & 68 & 62 & 66 \\
Hierarchical & 68 & -- & 68 & -- \\

\midrule
\rowcolor{gray!10}
\multicolumn{5}{c}{\textbf{\textit{MedQA}}} \\
\midrule
Hierarchical & 58 & -- & 58 & -- \\
Question-only & 65 & 62 & 63 & 65 \\
Tools + Roles & 64 & -- & 64 & -- \\

\midrule
\rowcolor{gray!10}
\multicolumn{5}{c}{\textbf{\textit{Musique}}} \\
\midrule
Question-only & 18 & 12 & 16 & 15 \\
Roles & 39 & 28 & 44 & 51 \\
Tools & 53 & 48 & 50 & 53 \\

\midrule
\rowcolor{gray!10}
\multicolumn{5}{c}{\textbf{\textit{Ethics}}} \\
\midrule
Debate & 66 & 57 & 60 & 63 \\
Roles & 71 & 59 & 60 & 56 \\

\midrule
\rowcolor{gray!10}
\multicolumn{5}{c}{\textbf{\textit{LiveCodeAgent}}} \\
\midrule
Hierarchical & 17 & -- & -- & -- \\
Homogeneous & 8 & 5 & 6 & -- \\
Roles & 7 & 14 & 10 & -- \\

\bottomrule
\end{tabular}
\caption{\textbf{Run-level accuracy by dataset and agent configuration.} Values report accuracy (\%). Column headers summarize the default agent compositions from the run summary. Role-based rows use task-specific agent roles with the corresponding base, alternate, expanded, or seven-agent role configurations.}
\label{tab:accuracy_by_agent_configuration}
\end{table*}

\section{Round Analysis}
\begin{figure*}[!htbp]
  \centering
  \includegraphics[width=\textwidth]{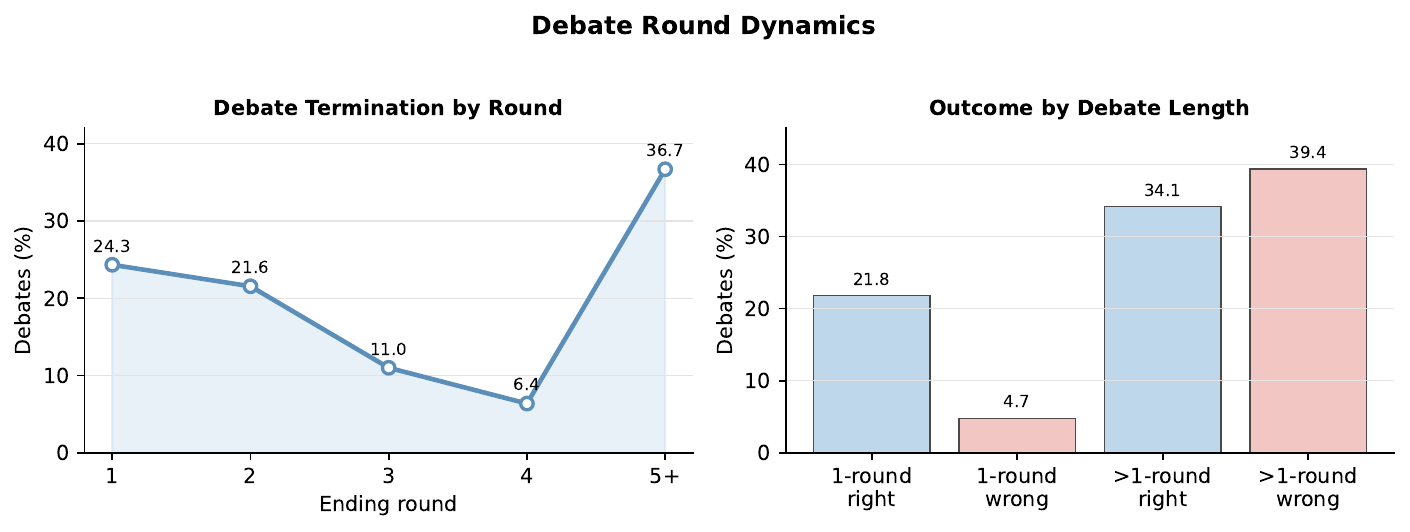}
  \caption{\textbf{Debate round dynamics.} Left: percentage of debates terminating at each round. Right: final correctness by debate length. One-round consensus is usually reliable, while multi-round debates account for most failures despite also producing many correct answers.}
  \label{fig:02debate_round_dynamics}
\end{figure*}

We analyze when debates terminate and how debate length relates to final correctness. Figure~\ref{fig:02debate_round_dynamics} shows that only 24.3\% of debates end after one round, while 75.7\% require additional deliberation and 36.7\% continue for five or more rounds. Less than half of debates end by Round~2, and a majority continue to at least Round~3, showing that long deliberation is common rather than exceptional.

Debate length is strongly associated with correctness. One-round consensus is relatively reliable: one-round traces account for 21.8\% of all correct outcomes and 4.7\% of all wrong outcomes, giving
$P(\text{correct} \mid \text{1 round}) = 21.8/(21.8+4.7) = 82.1\%$.
This suggests that immediate agreement often corresponds to easier or clearer instances. By contrast, multi-round debate is much riskier: multi-round traces account for 34.1\% of all correct outcomes and 39.4\% of all wrong outcomes, giving
$P(\text{correct} \mid >\text{1 round}) = 34.1/(34.1+39.4) = 46.4\%$.
Nearly all wrong answers come from this regime:
$P(>\text{1 round} \mid \text{wrong}) = 39.4/(39.4+4.7) = 89.2\%$.

This does not mean that additional rounds are useless. Multi-round traces also produce most correct answers:
$P(>\text{1 round} \mid \text{correct}) = 34.1/(21.8+34.1) = 61.0\%$.
The better interpretation is that multi-round debate is high-variance. It can support recovery, but it also contains most failures, making debate length a useful process-level signal for disagreement, answer drift, or failed correction.

\section{Taxonomy Saturation}
\label{app:taxonomy_saturation}

We analyze how quickly the taxonomy is recovered as more failed traces are sampled, treating each \texttt{unique\_identifier} as one trace. The full taxonomy contains 6 parent categories, 11 child categories, and 26 mapped leaf labels across 3,211 failed trajectories. For each sample size, we run 10 random trials, except for the full 3,211-trace setting, which is deterministic.

\begin{figure}[!htbp]
  \centering
  \includegraphics[width=\linewidth]{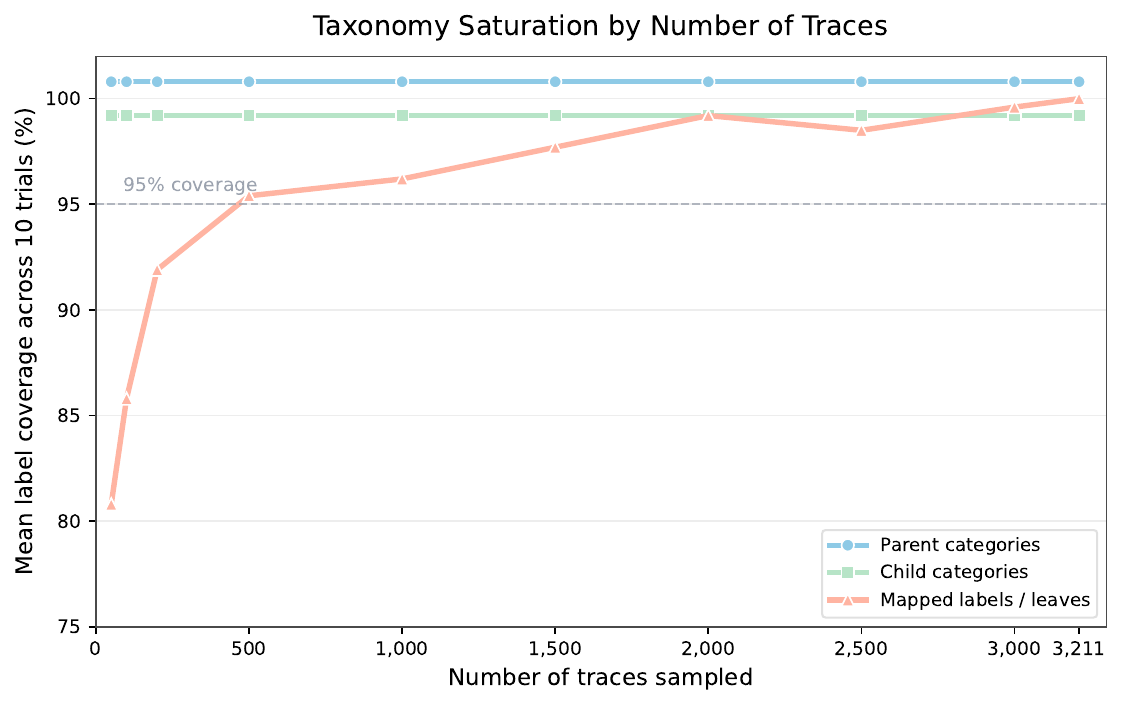}
  \caption{\textbf{Taxonomy saturation curve.} Parent and child categories saturate quickly, while fine-grained leaf labels require larger samples because several leaf errors are rare.}
  \label{fig:taxonomy_saturation}
\end{figure}

Figure~\ref{fig:taxonomy_saturation} shows that the high-level taxonomy saturates quickly. With only 50 sampled traces, we recover all parent and child categories. Fine-grained leaves saturate more slowly: 50 traces recover 80.8\% of leaves, 200 traces recover 91.9\%, and 500 traces recover 95.4\%. Near-complete leaf coverage requires much larger samples, with 2,000 traces recovering 99.2\% and the full dataset recovering all 26 leaves.

This pattern suggests that the broad structure of multi-agent failure is relatively stable, while rare fine-grained labels require substantially more data. The slowest labels to recover are extremely rare, including sycophancy, dogmatic disagreement or refusal to revise, tool/evidence output misread, poor information sharing, and false equivalence between distinct answers. Thus, small samples are sufficient for a usable parent/child taxonomy, but large-scale annotation is needed to capture rare leaf-level failure modes.

\section{Self-Reported Confidence Dynamics}
\label{app:confidence_dynamics}

Following prior work on verbalized or self-reported confidence in language models
\citep{lin2022teaching,tian-etal-2023-just,yona-etal-2024-large}, we analyze the confidence scores agents report during deliberation. We ask whether confidence tracks correctness, or whether deliberation itself increases confidence even when the group is wrong.

\begin{table*}[!t]
\centering
\scriptsize
\setlength{\tabcolsep}{3.5pt}
\renewcommand{\arraystretch}{1.08}
\begin{tabular}{@{}lcccl@{}}
\toprule
\textbf{Quantity} &
\textbf{Correct} &
\textbf{Incorrect} &
\textbf{$p$-value} &
\textbf{Interpretation} \\
\midrule

\makecell[l]{Confidence gain\\from first to final round}
& +5.54
& +10.49
& $5.30{\times}10^{-53}$
& \makecell[l]{Wrong traces gain\\more confidence} \\

\makecell[l]{Confidence gain\\after answer flip}
& +10.56
& +7.42
& $1.79{\times}10^{-16}$
& \makecell[l]{Flips increase confidence\\in both groups} \\

\makecell[l]{Final-round\\confidence}
& 93.10
& 89.34
& $3.94{\times}10^{-40}$
& \makecell[l]{Correct is higher,\\but wrong remains high} \\

\bottomrule
\end{tabular}
\caption{\textbf{Self-reported confidence dynamics.}
The first two rows report average confidence changes; the final row reports average final-round confidence. $p$-values compare correct and incorrect traces for each quantity.}
\label{tab:confidence_summary}
\end{table*}

Table~\ref{tab:confidence_summary} shows that confidence increases over deliberation, but not in a way that cleanly tracks correctness. Confidence rises for both correct and incorrect traces, with a larger increase for incorrect traces (+10.49) than correct traces (+5.54). This suggests that debate can amplify commitment to an answer even when the final answer is wrong, rather than simply reflecting progress toward the truth.

Answer changes are also confidence-amplifying. After an agent flips its answer, confidence increases in both finally correct and finally incorrect traces. The larger increase in correct traces suggests that some flips correspond to productive correction, but the positive increase for incorrect traces shows that agents can also become more confident after moving toward a wrong answer. Finally, correct traces end with higher confidence than incorrect traces, but the gap is modest: incorrect traces still end near 90\% confidence. Thus, we show that LLMs are not good at determining their confidence and hence should not be used for self-calibration.

\section{UNKNOWN Answer Behavior}
\label{app:unknown_answer_behavior}

We analyze cases where an agent outputs \textsc{UNKNOWN} during deliberation, separating blank responses, reasoning without a final answer, and explicit format failures. We also measure whether an \textsc{UNKNOWN} answer is recovered in the next round, meaning the same agent later gives a determinate answer, or persists, meaning the agent remains in an \textsc{UNKNOWN} state.

\begin{table*}[!t]
\centering
\scriptsize
\setlength{\tabcolsep}{2.6pt}
\renewcommand{\arraystretch}{1.08}
\begin{tabular}{@{}lcccccccl@{}}
\toprule
\textbf{Dataset} &
\makecell{\textbf{Debates w/}\\\textbf{UNKNOWN (\%)}} &
\makecell{\textbf{Dominant}\\\textbf{UNKNOWN type}} &
\makecell{\textbf{Blank}\\\textbf{(\%)}} &
\makecell{\textbf{Reasoning but}\\\textbf{UNKNOWN (\%)}} &
\makecell{\textbf{Format fail}\\\textbf{(\%)}} &
\makecell{\textbf{Recovery}\\\textbf{(\%)}} &
\makecell{\textbf{Persistence}\\\textbf{(\%)}} &
\textbf{Main pattern} \\
\midrule
Ethics   & $42.8 \pm 26.0$ & Blank response         & 43.0 & 23.5 & 4.5 & 52.0 & 48.0 & Mixed; often blank-driven \\
FinQA    & $25.0 \pm 16.3$ & Blank response          & 54.2 & 31.7 & 2.1 & 75.0 & 25.0 & UNKNOWN often resolves \\
HotpotQA & $39.9 \pm 29.7$ & Blank response        & 49.8 & 46.0 & 1.7 & 60.7 & 39.3 & Mixed but recoverable \\
MedQA    & $49.8 \pm 18.6$ & Reasoning without answer  & 10.2 & 75.8 & 9.9 & 10.7 & 89.3 & Highly persistent \\
MMLU     & $42.7 \pm 29.7$ & Reasoning without answer & 19.6 & 78.4 & 1.2 & 63.6 & 36.4 & Reasoned uncertainty often resolves \\
MuSiQue  & $67.2 \pm 32.7$ & Reasoning without answer & 46.6 & 48.0 & 0.8 & 40.1 & 60.0 & Frequent and often persistent \\
\midrule
\textbf{Overall} & $\mathbf{46.3 \pm 29.2}$ & Mixed & \textbf{36.9} & \textbf{52.7} & \textbf{3.0} & \textbf{50.4} & \textbf{49.6} & Mixed \\
\bottomrule
\end{tabular}
\caption{\textbf{\textsc{UNKNOWN} answer behavior by dataset.}
Blank responses and reasoning-but-\textsc{UNKNOWN} responses account for most \textsc{UNKNOWN} cases, while explicit format failures are relatively rare. Recovery and persistence measure whether an agent that produced \textsc{UNKNOWN} in one round gave a determinate answer or remained \textsc{UNKNOWN} in the next round.}
\label{tab:unknown_answer_behavior}
\end{table*}

Table~\ref{tab:unknown_answer_behavior} shows that \textsc{UNKNOWN} behavior is common but not uniform across datasets. Overall, 46.3\% of debates contain at least one \textsc{UNKNOWN} response. Most cases are not explicit format failures; they are either blank responses or responses that provide reasoning but do not commit to a final answer. This distinction matters because blank-driven \textsc{UNKNOWN}s may reflect generation or formatting failures, while reasoning-but-\textsc{UNKNOWN} responses are closer to unresolved uncertainty.

MedQA is the clearest persistent-\textsc{UNKNOWN} setting. Nearly half of MedQA debates contain \textsc{UNKNOWN}, and once an agent enters this state, it recovers in the next round only 10.7\% of the time, with 89.3\% persistence. MuSiQue also shows frequent and often persistent \textsc{UNKNOWN} behavior. By contrast, FinQA and HotpotQA show higher recovery, suggesting that some \textsc{UNKNOWN} states can be resolved when the task or setup provides enough evidence for later commitment. Ethics and FinQA have more blank-response behavior, so some \textsc{UNKNOWN}s in these datasets may reflect output failures rather than genuine epistemic uncertainty. Across settings, question-only setups appear most vulnerable to persistent \textsc{UNKNOWN} behavior, while tool or role structure is more likely to support recovery.

\pagebreak
\section{Description of taxonomy (detail)}

\begin{tcolorbox}[
  enhanced, breakable,
  title={Failure Taxonomy: Parent and Child Error Definitions},
  colback=gray!3, colframe=black!55, coltitle=black,
  fonttitle=\bfseries, boxrule=0.6pt, arc=1.5mm,
  left=1.2mm, right=1.2mm, top=1mm, bottom=1mm]
\footnotesize
\textbf{How to read the taxonomy.} Level-0 labels are broad parent error categories. Level-1 labels are the main error subtypes within each parent. Level-2 child labels are finer mechanisms used only where the data supported an additional split. Entries with no level-2 child label should be read at the level-1 definition.

\begin{description}[leftmargin=0.7cm, style=nextline, labelsep=0.2cm]
\item[\textbf{Recovery / correction failure}] After an error, disagreement, or correction opportunity appears, the group fails to detect, repair, or preserve the better-supported answer. \emph{(3,046 traces; 5,098 assignments).}
\begin{itemize}[leftmargin=0.45cm, itemsep=0.18em, topsep=0.15em]
\item \textbf{Persistent wrong framing after correction.} After a correction or competing framing appears, agents continue using the original flawed interpretation or reasoning frame. \emph{(1,993 traces; 2,616 assignments).}
\begin{itemize}[leftmargin=0.45cm, itemsep=0.08em, topsep=0.08em]
\item \emph{Anchor-driven recovery failure.} The group stays attached to an earlier wrong frame or answer and fails to recover the better-supported alternative after a correction opportunity appears. \emph{(1,067 traces; 1,213 assignments).}
\item \emph{Ignored or weakly adopted correction.} A correction is present but not incorporated strongly enough to change the trajectory of the final answer or reasoning. \emph{(697 traces; 767 assignments).}
\item \emph{Correction drift or oscillation.} The group partially corrects itself, but the fix is unstable, inconsistent, or gets reversed in later rounds. \emph{(379 traces; 395 assignments).}
\item \emph{Superficial patching without validation.} A correction is applied cosmetically or locally without checking whether it actually fixes the underlying error. \emph{(222 traces; 241 assignments).}
\end{itemize}
\item \textbf{Couldn't recover after unknown.} One or more agents output UNKNOWN or abstain, and the group fails to recover a substantive answer afterward. \emph{(739 traces; 763 assignments).}
\item \textbf{Failed recovery after disagreement.} Agents notice disagreement but do not adjudicate it successfully or turn it into a correct final answer. \emph{(631 traces; 643 assignments).}
\item \textbf{Failed recalculation or computational repair.} Agents attempt to fix a calculation or computational answer, but the repair remains wrong. \emph{(537 traces; 580 assignments).}
\item \textbf{Corrective signal ignored.} A useful correction, contradiction, test result, or warning appears, but agents do not incorporate it. \emph{(494 traces; 496 assignments).}
\end{itemize}
\item[\textbf{Consensus / decision failure}] The group converges, aggregates, or finalizes answers in a flawed way, such as drifting to a wrong majority or stabilizing too early. \emph{(2,677 traces; 4,147 assignments).}
\begin{itemize}[leftmargin=0.45cm, itemsep=0.18em, topsep=0.15em]
\item \textbf{Premature consensus.} Agents converge too early or follow agreement or majority without independently verifying whether the answer is correct. \emph{(1,913 traces; 2,272 assignments).}
\begin{itemize}[leftmargin=0.45cm, itemsep=0.08em, topsep=0.08em]
\item \emph{Social convergence without verification.} Agents agree because others agree, confidence rises through reinforcement, or consensus itself is treated as evidence rather than independently checked. \emph{(1,053 traces; 1,151 assignments).}
\item \emph{Anchor-driven lock-in.} An early answer, option, estimate, implementation, or dominant advocate fixes the group trajectory, and later rounds fail to escape it. \emph{(571 traces; 592 assignments).}
\item \emph{Premature closure on uncertain or incomplete evidence.} Agents stop too soon despite missing evidence, insufficient support, unknown or cannot-determine issues, unverified code, or incomplete validation. \emph{(367 traces; 376 assignments).}
\item \emph{Unresolved conflict masked by agreement.} The group appears to converge, but a real disagreement, inconsistency, edge case, or competing interpretation was never resolved. \emph{(153 traces; 153 assignments).}
\end{itemize}
\item \textbf{Oscillating consensus / answer instability.} The group repeatedly switches or splits among answers across rounds without stable evidence-based convergence. \emph{(834 traces; 961 assignments).}
\item \textbf{Answer drift.} The answer changes over rounds without a clear evidential reason or stable resolution. \emph{(790 traces; 859 assignments).}
\item \textbf{False equivalence between distinct answers.} The group treats different answers as equivalent when they refer to meaningfully different entities, values, dates, or claims. \emph{(55 traces; 55 assignments).}
\end{itemize}
\item[\textbf{Coordination / orchestration failure}] The debate structure, roles, moderator, turn-taking, state tracking, or protocol causes or amplifies the failure. \emph{(1,702 traces; 1,860 assignments).}
\begin{itemize}[leftmargin=0.45cm, itemsep=0.18em, topsep=0.15em]
\item \textbf{Moderator or hierarchy failure.} A moderator, supervisor, hierarchy, or aggregation step selects poorly, fails to resolve conflict, or amplifies an error. \emph{(1,379 traces; 1,456 assignments).}
\begin{itemize}[leftmargin=0.45cm, itemsep=0.08em, topsep=0.08em]
\item \emph{Unstable coordination.} The group drifts, flips, oscillates, churns, or fails to maintain a stable shared direction across agents or rounds. \emph{(856 traces; 884 assignments).}
\item \emph{Weak oversight or improper question asking between rounds.} The moderator or supervisor fails to ask the right follow-up questions, enforce verification, direct agents toward unresolved issues, integrate partial fixes, or preserve corrections between rounds. \emph{(348 traces; 351 assignments).}
\item \emph{Failed aggregation or final selection.} The hierarchy, moderator, supervisor, or final synthesis step selects, formats, aggregates, or outputs the wrong final answer despite available agent signals. \emph{(219 traces; 221 assignments).}
\end{itemize}
\item \textbf{Role/access coordination failure.} Assigned roles, permissions, context access, or division of labor cause missing information or poor coordination. \emph{(393 traces; 404 assignments).}
\end{itemize}
\item[\textbf{Tool / evidence-use failure}] Agents fail to retrieve, interpret, share, verify, or use external evidence or tool outputs correctly. \emph{(1,430 traces; 1,709 assignments).}
\begin{itemize}[leftmargin=0.45cm, itemsep=0.18em, topsep=0.15em]
\item \textbf{Evidence retrieval or mapping failure.} Agents fail to retrieve the needed evidence or map retrieved evidence to the correct entity, relation, step, or question. \emph{(715 traces; 780 assignments).}
\item \textbf{Weak, fabricated, or unsupported evidence grounding.} The answer is justified with weak evidence, fabricated support, hallucinated facts, or evidence that does not actually ground the claim. \emph{(567 traces; 615 assignments).}
\item \textbf{Tool/evidence result ignored or not shared.} A relevant tool or evidence result is available but ignored, omitted from the debate, or not shared with agents who need it. \emph{(300 traces; 300 assignments).}
\item \textbf{Tool/evidence output misread.} A tool result, retrieved passage, or evidence source is available but is interpreted incorrectly. \emph{(14 traces; 14 assignments).}
\end{itemize}
\item[\textbf{Reasoning / task-understanding failure}] An agent misunderstands the task, uses an incorrect framing, makes an invalid inference, or gives reasoning that does not support its answer. \emph{(1,390 traces; 1,789 assignments).}
\begin{itemize}[leftmargin=0.45cm, itemsep=0.18em, topsep=0.15em]
\item \textbf{Unsupported assumption or guessing.} Agents assert an answer, bridge, fact, or inference without adequate evidence, often relying on plausible-sounding guesses. \emph{(764 traces; 855 assignments).}
\item \textbf{Question misinterpretation.} Agents misunderstand what the question is asking, including the target entity, relation, constraint, or wording. \emph{(351 traces; 370 assignments).}
\item \textbf{Edge-case handling failure.} The solution fails because agents miss an edge case, exception, boundary condition, or special-case requirement. \emph{(249 traces; 293 assignments).}
\item \textbf{Calculation, rounding, precision, or representation error.} The failure comes from arithmetic, numeric precision, rounding, units, formatting of quantities, or representation of a computed result. \emph{(149 traces; 154 assignments).}
\item \textbf{Output-format or instruction-following error.} The response violates required format, answer extraction rules, tool protocol, or other explicit instructions. \emph{(114 traces; 117 assignments).}
\end{itemize}
\item[\textbf{Communication / deliberation failure}] Agents fail to exchange, challenge, interpret, or respond to each other's reasoning effectively. \emph{(456 traces; 484 assignments).}
\begin{itemize}[leftmargin=0.45cm, itemsep=0.18em, topsep=0.15em]
\item \textbf{Low-signal contribution or non-participation.} An agent gives little useful content, repeats others, abstains, outputs blank or UNKNOWN, or does not contribute to resolving the task. \emph{(186 traces; 189 assignments).}
\item \textbf{Ignored or weakly challenged evidence/reasoning.} Weak, unsupported, or conflicting reasoning is accepted without adequate scrutiny, or strong evidence is not challenged or used well. \emph{(156 traces; 161 assignments).}
\item \textbf{Interpretation mismatch between agents.} Agents operate under different interpretations of the task, question, terms, or answer target. \emph{(98 traces; 99 assignments).}
\item \textbf{Poor information sharing.} Relevant information is held by one agent or part of the system but is not communicated clearly to others. \emph{(29 traces; 29 assignments).}
\item \textbf{Dogmatic disagreement or refusal to revise.} Agents hold onto incompatible positions despite counterarguments or evidence, preventing resolution. \emph{(5 traces; 5 assignments).}
\item \textbf{Sycophancy.} Agents defer to, flatter, or agree with another agent, user, or majority instead of independently evaluating the claim. \emph{(1 trace; 1 assignment).}
\end{itemize}
\end{description}
\end{tcolorbox}

\pagebreak
\section{Prompts for all setup}


\begin{tcolorbox}[
  enhanced, breakable,
  title={Trace Prompt 1: Question Only Initial Answer},
  colback=red!5,
colframe=red!35,coltitle=black,
  fonttitle=\bfseries, boxrule=0.6pt, arc=1.5mm]
\footnotesize
\textbf{System prompt.} You are a dataset-specific debate agent answering from the question text alone. Use careful reasoning, but do not invent unsupported details. Keep the response concise.

\textbf{User prompt.} Answer the given question with the shortest answer span you can support. Do not use external context. You may write brief reasoning first if needed. End with two final lines: one line giving the final answer only, and one line giving confidence as a number from 0 to 100. If unsure, still choose the best short answer span.
\end{tcolorbox}

\begin{tcolorbox}[
  enhanced, breakable,
  title={Trace Prompt 2: Shared Context or Tool Access Initial Answer},
  colback=red!5,
colframe=red!35, coltitle=black,
  fonttitle=\bfseries, boxrule=0.6pt, arc=1.5mm]
\footnotesize
\textbf{System prompt.} You are a dataset-specific debate agent with access to supporting context or tool output. Use the provided evidence carefully and answer with the shortest correct answer span. Keep the response concise.

\textbf{User prompt.} Answer the given question using the supplied context or tool output as evidence. You may write brief reasoning first if needed. End with two final lines: one line giving the final answer only, and one line giving confidence as a number from 0 to 100. Do not add explanation after the final answer.
\end{tcolorbox}

\begin{tcolorbox}[
  enhanced, breakable,
  title={Trace Prompt 3: Role Based Debate Setup},
  colback=red!5,
colframe=red!35,coltitle=black,
  fonttitle=\bfseries, boxrule=0.6pt, arc=1.5mm]
\footnotesize
\textbf{System prompt.} Each agent receives a role. The evidence reader can use supporting context and should extract the exact answer span. The skeptic challenges weak reasoning and searches for inconsistencies. The verifier checks whether proposed answers fit the question and debate history. The synthesizer combines the debate into one concise answer.

\textbf{User prompt.} Answer the question according to your assigned role. If you have context access, use the supporting context. If you do not have context access, rely on your role and the debate history. End with the final answer only and a confidence score from 0 to 100.
\end{tcolorbox}

\begin{tcolorbox}[
  enhanced, breakable,
  title={Trace Prompt 4: Follow Up Debate Round},
  colback=red!5,
colframe=red!35, coltitle=black,
  fonttitle=\bfseries, boxrule=0.6pt, arc=1.5mm]
\footnotesize
\textbf{System prompt.} You are now participating in a multi-agent debate. Keep the response concise and stay within your assigned role.

\textbf{User prompt.} Review your previous answer, your previous reasoning, and the other agents' answers, confidence scores, and reasoning from the prior round. Briefly evaluate whether the other agents provide stronger evidence or reasoning than you did. State whether you keep or change your answer. End with the final answer only and a confidence score from 0 to 100.
\end{tcolorbox}

\begin{tcolorbox}[
  enhanced, breakable,
  title={Trace Prompt 5: Consensus Judge},
  colback=red!5,
colframe=red!35, coltitle=black,
  fonttitle=\bfseries, boxrule=0.6pt, arc=1.5mm]
\footnotesize
\textbf{System prompt.} You are a strict consensus judge. Determine whether the agents have reached consensus on the same answer. Only mark consensus when all answers are equivalent in the context of the question.

\textbf{User prompt.} Given the question and the current answers from all agents, decide whether the answers are effectively the same. Treat aliases, abbreviations, spelling variants, and equivalent phrasing as consensus. Do not mark consensus if any answer differs meaningfully, names a different entity, gives a different date or number, or only partially overlaps with another answer. Return a yes or no consensus decision and, if yes, the shortest clean canonical answer.
\end{tcolorbox}

\begin{tcolorbox}[
  enhanced, breakable,
  title={Trace Prompt 6: Moderator Final Decision},
  colback=red!5,
colframe=red!35, coltitle=black,
  fonttitle=\bfseries, boxrule=0.6pt, arc=1.5mm]
\footnotesize
\textbf{System prompt.} You are a careful moderator. Choose the most likely concise answer span when the agents do not reach consensus.

\textbf{User prompt.} Review the question and the agents' answers across all debate rounds. Choose the most likely concise final answer. If the debate contains equivalent paraphrases, choose the shortest clean answer. End with the final answer only and a confidence score from 0 to 100.
\end{tcolorbox}

\begin{tcolorbox}[
  enhanced, breakable,
  title={Taxonomy Prompt 1: Trajectory Card Extraction},
  colback=red!5,
colframe=red!35, coltitle=black,
  fonttitle=\bfseries, boxrule=0.6pt, arc=1.5mm]
\footnotesize
\textbf{System prompt.} You are generating a trajectory card from a multi-agent debate trace. Focus on communication, coordination, orchestration, tool use, and recovery failures. Distinguish agent-level trajectories, round-progression errors, and whole-trajectory errors.

\textbf{User prompt.} Read the trace and produce a concise trajectory card. Describe how each agent's trajectory contributed to failure, including answer shifts, agreement patterns, tool use, and failed corrections. Identify what went wrong from one round to the next. Identify the overall cross-round failure pattern. For every item, include the concrete round or step and the responsible agents. End with a one-sentence summary of why the trajectory failed.
\end{tcolorbox}

\begin{tcolorbox}[
  enhanced, breakable,
  title={Taxonomy Prompt 2: Candidate Failure Tag Generation},
  colback=red!5,
colframe=red!35, coltitle=black,
  fonttitle=\bfseries, boxrule=0.6pt, arc=1.5mm]
\footnotesize
\textbf{System prompt.} You are generating reusable failure tags for a trajectory card. Choose concise labels that explain the main failure modes.

\textbf{User prompt.} Using the trace and trajectory-card summary, produce several concise candidate failure tags. Each tag should be grounded in evidence from specific rounds, steps, or agents. Prefer reusable labels such as premature consensus, ignored evidence, weak correction, poor persuasion, answer drift, overconfidence, bad coordination, failed recovery, tool misuse, and orchestration failure. For each tag, include a short justification and evidence reference.
\end{tcolorbox}

\begin{tcolorbox}[
  enhanced, breakable,
  title={Taxonomy Prompt 3: Broad Parent Category Mapping},
  colback=red!5,
colframe=red!35, coltitle=black,
  fonttitle=\bfseries, boxrule=0.6pt, arc=1.5mm]
\footnotesize
\textbf{System prompt.} You are a careful taxonomy annotator.

\textbf{User prompt.} Categorize one multi-agent failure tag into exactly one broad parent category. Use the tag, evidence, justification, and trace identifier to disambiguate. Choose from reasoning or task-understanding failure, communication or deliberation failure, consensus or decision failure, recovery or correction failure, coordination or orchestration failure, and tool or evidence-use failure. If the tag does not naturally fit any category, mark it as unmapped. Do not invent new broad categories. Return the chosen category and a one-sentence reason.
\end{tcolorbox}

\begin{tcolorbox}[
  enhanced, breakable,
  title={Taxonomy Prompt 4: Fine Grained Label Mapping},
  colback=red!5,
colframe=red!35, coltitle=black,
  fonttitle=\bfseries, boxrule=0.6pt, arc=1.5mm]
\footnotesize
\textbf{System prompt.} You are a careful taxonomy annotator.

\textbf{User prompt.} Map one taxonomy row to exactly one fine-grained label from the allowed label list. Use only the row fields: broad category, representative tag, representative evidence, and representative justification. Choose the primary failure mode, not a secondary detail. If the row does not clearly match any allowed label, mark it as unmapped rather than forcing a poor fit. Return the chosen label and a one-sentence reason.
\end{tcolorbox}

\begin{tcolorbox}[
  enhanced, breakable,
  title={Taxonomy Prompt 5: Micro Cluster Review},
  colback=red!5,
colframe=red!35, coltitle=black,
  fonttitle=\bfseries, boxrule=0.6pt, arc=1.5mm]
\footnotesize
\textbf{System prompt.} You are a careful taxonomy-building assistant.

\textbf{User prompt.} Review one semantic micro-cluster of currently unmapped taxonomy rows. The rows were clustered using representative tags. Decide whether the cluster is coherent. Prefer mapping to an existing label if one fits. Propose a new label only when the cluster is coherent and not covered by existing labels. If the cluster is mixed, vague, or too small to support a label, keep it unmapped. Return the decision, chosen label, one-sentence label definition, rationale, subthemes, confidence, and a note for manual review.
\end{tcolorbox}

\begin{tcolorbox}[
  enhanced, breakable,
  title={Taxonomy Prompt 6: Cluster Decision Consolidation},
  colback=red!5,
colframe=red!35, coltitle=black,
  fonttitle=\bfseries, boxrule=0.6pt, arc=1.5mm]
\footnotesize
\textbf{System prompt.} You are a careful taxonomy consolidation assistant.

\textbf{User prompt.} Consolidate the cluster-level taxonomy decisions for one parent category. Merge repeated or synonymous proposed labels. Prefer existing labels when a proposed label is really a subtype of an existing label. Create a final new label only if multiple clusters support a genuinely distinct pattern. Do not preserve tiny one-off labels as final taxonomy labels. For each cluster, provide the final decision, final label, and brief reason. Also summarize the consolidated categories and overall notes.
\end{tcolorbox}

\begin{tcolorbox}[
  enhanced, breakable,
  title={Taxonomy Prompt 7: Level Two Child Label Assignment},
  colback=red!5,
colframe=red!35, coltitle=black,
  fonttitle=\bfseries, boxrule=0.6pt, arc=1.5mm]
\footnotesize
\textbf{System prompt.} You are a careful taxonomy annotator. Classify every row into exactly one allowed child label.

\textbf{User prompt.} Given a parent broad category, a parent label, allowed child labels, and a batch of rows, assign one child label to every row. Do not use unmapped, unknown, other, mixed, or manual-review labels. Choose the dominant failure mechanism in the representative tag and justification. For premature consensus, prefer social convergence without verification when agreement or peer reinforcement is central; prefer anchor-driven lock-in when an early answer or advocate controls later rounds; prefer unresolved conflict masked by agreement when consensus hides a real disagreement; and prefer premature closure on incomplete evidence when agents stop too soon despite missing support. Return each row's child label, rationale, and confidence.
\end{tcolorbox}